\documentclass[runningheads]{llncs}
\usepackage{graphicx}
\usepackage{tikz}
\usepackage{comment}
\usepackage{amsmath,amssymb} % define this before the line numbering.
\usepackage{color}
\usepackage{cite}
\usepackage{tabu}  

\usepackage[pagebackref=true,breaklinks=true,letterpaper=true,colorlinks,bookmarks=false]{hyperref}
\newcommand\blfootnote[1]{%
  \begingroup
  \renewcommand\thefootnote{}\footnote{#1}%
  \addtocounter{footnote}{-1}%
  \endgroup
}

\begin{document}
% \renewcommand\thelinenumber{\color[rgb]{0.2,0.5,0.8}\normalfont\sffamily\scriptsize\arabic{linenumber}\color[rgb]{0,0,0}}
% \renewcommand\makeLineNumber {\hss\thelinenumber\ \hspace{6mm} \rlap{\hskip\textwidth\ \hspace{6.5mm}\thelinenumber}}
% \linenumbers
\pagestyle{headings}
\mainmatter
\def\ECCVSubNumber{100}  % Insert your submission number here

\title{AIM 2020: Scene Relighting and Illumination Estimation Challenge} % Replace with your title

% INITIAL SUBMISSION 
\begin{comment}
\titlerunning{ECCV-20 submission ID \ECCVSubNumber} 
\authorrunning{ECCV-20 submission ID \ECCVSubNumber} 
\author{Anonymous ECCV submission}
\institute{Paper ID \ECCVSubNumber}
\end{comment}
%******************

% CAMERA READY SUBMISSION
% \begin{comment}
\titlerunning{AIM 2020 Relighting Challenge}
% If the paper title is too long for the running head, you can set
% an abbreviated paper title here
%
\index{El Helou, Majed}
\author{Majed El Helou\inst{1} \and
Ruofan Zhou\inst{1} \and
Sabine S\"usstrunk\inst{1} \and
Radu Timofte\inst{2} \and
Mahmoud Afifi$^*$\and
Michael S. Brown$^*$\and
Kele Xu$^*$ \and
Hengxing Cai$^*$\and
Yuzhong Liu$^*$\and
Li-Wen Wang$^*$\and
Zhi-Song Liu$^*$\and
Chu-Tak Li$^*$\and
Sourya Dipta Das$^*$\and 
Nisarg A. Shah$^*$\and
Akashdeep Jassal$^*$\and
Tongtong Zhao$^*$\and
Shanshan Zhao$^*$\and
Sabari Nathan$^*$\and
M. Parisa Beham$^*$\and
R. Suganya$^*$\and
Qing Wang$^*$\and
Zhongyun Hu$^*$\and
Xin Huang$^*$\and
Yaning Li$^*$\and
Maitreya Suin$^*$\and
Kuldeep Purohit$^*$\and
A. N. Rajagopalan$^*$\and
Densen Puthussery$^*$\and
Hrishikesh P S$^*$\and
Melvin Kuriakose$^*$\and
Jiji C V$^*$\and
Yu Zhu$^*$\and
Liping Dong$^*$\and
Zhuolong Jiang$^*$\and
Chenghua Li$^*$\and
Cong Leng$^*$\and
Jian Cheng$^*$
}

% \author{First Author\inst{1}\orcidID{0000-1111-2222-3333} \and
% Second Author\inst{2,3}\orcidID{1111-2222-3333-4444} \and
% Third Author\inst{3}\orcidID{2222--3333-4444-5555}}
\authorrunning{M. El Helou, R. Zhou, S. S\"usstrunk, R. Timofte \textit{et al.}}

\institute{EPFL, Switzerland \and
ETHZ, Switzerland}
% \textcolor{red}{Team member affiliations are listed in the appendix, c.f. the footnote on this page, please do not repeat them here, it would become too cumbersome.} \and
%  Jadavpur University, India\and
%  IIT Jodhpur, India\and
%  PEC, India \and
%  DMU, China \and
%  CEB, China\and
%  Couger Inc,Japan \and
%  SIT,India \and
%  TCE,India}
% \email{lncs@springer.com}\\
% \url{http://www.springer.com/gp/computer-science/lncs} \and
% ABC Institute, Rupert-Karls-University Heidelberg, Heidelberg, Germany\\
% \email{\{abc,lncs\}@uni-heidelberg.de}}
% \end{comment}
%******************
\maketitle

\begin{abstract}
We review the AIM 2020 challenge on virtual image relighting and illumination estimation. This paper presents the novel VIDIT dataset used in the challenge and the different proposed solutions and final evaluation results over the 3 challenge tracks. The first track considered one-to-one relighting; the objective was to relight an input photo of a scene with a different color temperature and illuminant orientation (i.e., light source position). The goal of the second track was to estimate illumination settings, namely the color temperature and orientation, from a given image. Lastly, the third track dealt with any-to-any relighting, thus a generalization of the first track. The target color temperature and orientation, rather than being pre-determined, are instead given by a guide image. Participants were allowed to make use of their track 1 and 2 solutions for track 3. The tracks had 94, 52, and 56 registered participants, respectively, leading to 20 confirmed submissions in the final competition stage.
\end{abstract}

\blfootnote{Majed El Helou, Ruofan Zhou, Sabine S\"usstrunk \textit{(majed.elhelou, ruofan.zhou, sabine.susstrunk)@epfl.ch}, and Radu Timofte \textit{radu.timofte@vision.ee.ethz.ch}, are the challenge organizers, and the other authors are challenge participants.\\ $^*$Appendix~\ref{sec:teams} lists all the teams and affiliations.}

\section{Introduction}
~\blfootnote{\url{https://github.com/majedelhelou/VIDIT}}
Deep image relighting has multiple applications both in research and in practice, and is recently witnessing increased interest. A single-image relighting method would allow aesthetic enhancement applications, such as photo montage of images taken under different illuminations, and illumination retouching without human expert work. Very importantly, in computer vision research image relighting can be leveraged for data augmentation, enabling the trained methods to be robust to changes in light source position or color temperature. It could also serve for domain adaptation, by normalizing input images to a unique set of illumination settings that the down-stream computer vision method was trained on. The relighting task contains multiple sub-tasks, namely, illumination estimation and manipulation, shadow removal or practically inpainting for hardly lit areas, and geometric understanding for shadow recasting. The combination of these tasks makes relighting very challenging. 

Recently, datasets limited to interior scenes~\cite{murmann2019dataset}, underexposed images enhanced by professionals~\cite{wang2019underexposed}, and rendered images with randomized light directions~\cite{xu2018deep} have been proposed, but none serve the benchmarking needs for image relighting, namely, having all $M\times N$ combinations of $M$ scenes and $N$ illumination settings. 
Further datasets are used in the literature on style transfer or intrinsic image decomposition. For instance, IIW~\cite{bell2014intrinsic} and SAW~\cite{kovacs2017shading} contain human-labeled reflectance and shading annotations, and BigTime~\cite{li2018learning} contains time-lapse data of scenes illuminated under varying light conditions. Multiple methods are recently being developed for relighting~\cite{sun2019single,nagano2019deep,dherse2020scene}, and the prior literature on intrinsic images, which disentangle surface reflectance from lighting, is rich~\cite{weiss2001deriving,tappen2003recovering,finlayson2004intrinsic,shen2011intrinsic,barron2012color,bell2014intrinsic}, notably for applications such as relighting~\cite{bousseau2009user} and normalization~\cite{matsushita2004illumination}.

The aim of this challenge, and of the novel dataset \textbf{V}irtual \textbf{I}mage \textbf{D}ataset for \textbf{I}llumination \textbf{T}ransfer (VIDIT), is to gauge the current state-of-the-art for image relighting. The virtual dataset provides a well-controlled setup to provide full-reference evaluation, which is ideal for benchmarking purposes, and is an important step towards real-image relighting. Such virtual datasets have proven useful in multiple applications to augment even the training datasets containing real images, for instance the vKitti data~\cite{cabon2020virtual}. There could be differences relative to real images such as the distribution of textures that can vary from man-made to natural scenes~\cite{torralba2003statistics,burton1987color}, the specifics of the capturing device like chromatic aberrations~\cite{elhelou2018aam,llanos2020simultaneous,zhao2020modified}, or the presence of multiple light sources. VIDIT itself is described in the following section. The goal of the challenge is thus to provide a benchmark on this dataset for future research on image relighting.

This challenge is one of the AIM 2020 associated challenges on:
scene relighting and illumination estimation~\cite{elhelou2020aim_relighting}, image extreme inpainting~\cite{ntavelis2020aim_inpainting}, learned image signal processing pipeline~\cite{ignatov2020aim_ISP}, rendering realistic bokeh~\cite{ignatov2020aim_bokeh}, real image super-resolution~\cite{wei2020aim_realSR}, efficient super-resolution~\cite{zhang2020aim_efficientSR}, video temporal super-resolution~\cite{son2020aim_VTSR} and video extreme super-resolution~\cite{fuoli2020aim_VXSR}.

\section{Scene relighting and illumination estimation challenge}
\subsection{Dataset}
The challenge, whose 3 tracks are described in the following section, is based on a novel dataset: VIDIT~\cite{elhelou2020vidit}. VIDIT contains 300 virtual scenes used for training, where every scene is captured 40 times in total: from 8 equally-spaced azimuthal angles, each lit with 5 different illuminants. Every image is $1024\times 1024$, but the images are downsampled by a factor of 2, with bicubic interpolation over $4\times 4$ windows, to ease computations for track 3. The dataset is publicly available (\url{https://github.com/majedelhelou/VIDIT}).

\subsection{Tracks and competition}
\textbf{Track 1: One-to-one relighting.} \\
\textbf{Description:} the relighting task is pre-determined and fixed for all validation and test samples. In other words, the objective is to manipulate an input image from one pre-defined set of illumination settings (namely, North, 6500K) to another pre-defined set (East, 4500K). The images are in $1024\times 1024$ resolution, both input and output, and nothing other than the input image is provided. \\
\textbf{Evaluation protocol:} We evaluate the results using the PSNR and SSIM~\cite{wang2004image} metrics, and the self-reported run-times and implementation details are also provided. For the final ranking, we define a Mean Perceptual Score (MPS) as the average of the normalized SSIM and LPIPS~\cite{zhang2018unreasonable} scores, themselves averaged across the entire test set of each submission
\begin{equation}\label{eq:MPS}
    0.5\cdot(S + (1-L)),
\end{equation}
where $S$ is the SSIM score, and $L$ is the LPIPS score. We note that normalizing $S$ and $(1-L)$, by dividing them respectively by their maximum values across all the track's submissions, before averaging the two does not affect the final ranking. We thus do not do this normalization, which also makes it simpler for external comparisons. \\

\noindent \textbf{Track 2: Illumination settings estimation.} \\
\textbf{Description:} the goal of this track is to estimate, from a single input image, the illumination settings that were used in rendering it. Given the input image, the output should estimate the color temperature of the illuminant as well as the orientation, i.e. the position of the light source. The input images are also $1024\times 1024$ and no other input is given than the 2D image. \\
\textbf{Evaluation protocol:} The evaluation of track 2 is based on the accuracy of predictions following this formula for the loss
\begin{equation}\label{eq:track2_loss}
    \sqrt{ \sum_{i=0}^{N-1} \left( \frac{|\hat{\phi_i}-\phi_i|mod180}{180} \right)^2 + (\hat{T_i}-T_i)^2 }
\end{equation}
where $\hat{\phi_i}$ is the predicted angle (0-360) for test sample $i$ and $\phi_i$ is the ground-truth value for that sample. $\hat{T_i}$ is the temperature prediction for test sample $i$ and $T_i$ is the ground-truth value for that sample. $T_i$ takes values equal to [0, 0.25, 0.5, 0.75, 1], which correspond to the color temperature values [2500K, 3500K, 4500K, 5500K, 6500K]. \\

\noindent \textbf{Track 3: Any-to-any relighting.} \\
\textbf{Description:} this track is a generalization of the first track. The objective is to relight an input image (both color temperature and light source position manipulation) from any arbitrary illumination settings to any arbitrary illumination settings. The latter settings are dictated by a second input guide image, as in style transfer applications. The participants were allowed to make use of their solutions to the first two tracks to develop a solution for this track. The images are in $512\times 512$ resolution to ease computations, as this track is very challenging. \\
\textbf{Evaluation protocol:} We carry out a similar evaluation as for track 1. As the inputs are pairs of possible test images, they cover a larger span of candidate options. For that reason, we double the number of data samples in the validation and test sets for this track. \\

\noindent \textbf{Challenge phases for all tracks.} (1) Development: registered teams were given access to the training input and target data, as well as the input validation set data. An online validation server with a leader board provided automated feedback for the submitted image results on the validation set, which was made up of 45 images for tracks 1 and 2, and 90 image pairs for track 3; (2) Testing: registered teams were given access to the input test sets, which are of the same size as the validation ones, and could submit their test results to a private test server. For a submission to be accepted, open-source code and a fact sheet detailing the implemented method needed to be submitted along with the test results. Test results were kept hidden from participating teams, to avoid any chances of test over-fitting, and were only revealed at the end of the challenge.

\section{Challenge results}
The results of all three tracks are collected in Tables~\ref{table:T1},~\ref{table:T2}, and~\ref{table:T3}, respectively. The top solutions are described in the following sections, and the remainder is in the supplementary material.

Visual results of some top submissions along with input and ground-truth images for track 1 are shown in Fig.~\ref{fig:t1_visual}. We notice that most of the outputs generate the relit image with the correct color temperature, however, the shadows are harder to estimate. For instance, lyl and YorkU suffer from shadow removal. Both CET\_SP and CET\_CVLab tend to remove the unnecessary shadows, although not perfectly, which underlines the difficulty of the shadow-relighting sub-task. 
We show visual results of some submissions to track 3 in Fig.~\ref{fig:t3_visual}. Among the top 3 submissions, only NPU-CVPG is able to successfully relight the bottom-right part and produce the closest color temperature to the ground-truth. 

\begin{table}[htp!]
\centering
\resizebox{\textwidth}{!}{%
\begin{tabular}{l|l||c|c|c|c||c|c|c}
Team&Author &\textbf{MPS} $\uparrow$&SSIM $\uparrow$&LPIPS $\downarrow$&PSNR $\uparrow$&Run-time&Platform&GPU\\
\hline\hline
CET\_SP$^*$&hrishikeshps&0.6452 (-)&0.6310 (2)&0.3405 \textbf{(1)}&17.0717 (2)&0.03s&Tensorflow&P100\\
CET\_CVLab&Densen&0.6451 \textbf{(1)}&0.6362 \textbf{(1)}&0.3460 (3)&16.8927 (6)&0.03s&Tensorflow&P100\\
lyl&tongtong&0.6436 (2)&0.6301 (3)&0.3430 (2)&16.6801 (8)&13s&PyTorch&V100\\
YorkU&mafifi&0.6216 (3)&0.6091 (4)&0.3659 (5)&16.8196 (7)&6s&PyTorch&1080TI\\
IPCV\_IITM&ms\_icpv&0.5897 (4)&0.5298 (7)&0.3505 (4)&17.0594 (3)&0.04s&PyTorch&Titan X\\
DeepRelight&leven&0.5892 (5)&0.5928 (6)&0.4144 (7)&17.4252 \textbf{(1)}&0.5s&PyTorch&2080TI\\
Withdrawn&tomanut&0.5603 (6)&0.5236 (8)&0.4029 (6)&16.5136 (9)&0.01s&PyTorch&2080TI\\
Hertz&souryadipta&0.5339 (7)&0.5666 (6)&0.4989 (8)&16.9234 (4)&0.006s&PyTorch&1080TI\\
Image Lab&sabarinathan&0.3746 (8)&0.3769 (9)&0.6278 (9)&16.8949 (5)&0.12s&Tensorflow&1080TI\\
\hline\hline
input image & - & 0.6438 &0.6288 & 0.3412 & 16.2796
\end{tabular}}
\caption{AIM 2020 Image Relighting Challenge Track 1 (One-to-one relighting) results. The MPS, used to determine the final ranking, is computed following Eq.~\eqref{eq:MPS}. $^*$CET\_CVLab and CET\_SP are merged into one, due to large similarity between the proposed solutions.%, and for fairness to other teams we use the worse-performing out of their two solutions for the ranking (they nevertheless share the first place). 
We also note that normalizing SSIM and (1-LPIPS) scores by the maximum in the track, for computing the MPS, does not affect the ranking.}
\label{table:T1}
\end{table}

\begin{table}[htp!]
\centering
\resizebox{\textwidth}{!}{%
\begin{tabular}{l|l||c|c|c||c|c|c}
Team&Author &\textbf{Loss} $\downarrow$&AngLoss $\downarrow$&TempLoss $\downarrow$&Run-time&Platform&GPU\\
\hline\hline
AiRiA\_CG&Airia\_CG&0.0875 \textbf{(1)}&0.0722 (3)&0.0153 \textbf{(1)}&0.03s&PyTorch&Titan Xp\\
YorkU&mafifi&0.0887 (2)&0.0639 (2)&0.0248 (2)&0.95s&MATLAB&1080TI\\
Image Lab&sabarinathan&0.0984 (3)&0.0513 \textbf{(1)}&0.0471 (5)&0.02s&Tensorflow&1080TI\\
debut\_kele&debut\_kele&0.1431 (4)&0.1125 (4)&0.0306 (3)&&\\
RGETH&Georgechogovadze&0.1708 (5)&0.1347 (5)&0.0361 (4)&0.026s&PyTorch&\\
\hline\hline
random guess&-&0.5987&0.3729&0.2257\\
\end{tabular}}
\caption{AIM 2020 Image Relighting Challenge Track 2 (Illumination settings estimation) results. The loss is computed based on the angle and color temperature predictions, following Eq.~\eqref{eq:track2_loss}, and is used to determine the final ranking.}
\label{table:T2}
\end{table}

\begin{table}[htp!]
\centering
\resizebox{\textwidth}{!}{%
\begin{tabular}{l|l||c|c|c|c||c|c|c}
Team&Author &\textbf{MPS} $\uparrow$&SSIM $\uparrow$&LPIPS $\downarrow$&PSNR $\uparrow$&Run-time&Platform&GPU\\
\hline\hline
NPU-CVPG&walden&0.6484 \textbf{(1)}&0.6353 \textbf{(1)}&0.3386 (3)&18.5436 (2)&0.15s&PyTorch&1080TI\\
YorkU&mafifi&0.6428 (2)&0.6195 (2)&0.3338 (2)&18.2384 (4)&6s&PyTorch&1080TI\\
IPCV\_IITM&ms\_icpv&0.6424 (3)&0.6042 (3)&0.3194 \textbf{(1)}&19.3559 \textbf{(1)}&0.3s&PyTorch&Titan X\\
lyl&tongtong&0.6213 (4)&0.5881 (4)&0.3455 (4)&17.6314 (5)&13s&PyTorch&V100\\
AiRiA\_CG&Airia\_CG&0.5258 (5)&0.4451 (5)&0.3936 (5)&18.3493 (3)&&PyTorch&Titan Xp\\
RGETH&Georgechogovadze&0.3465 (6)&0.4123 (6)&0.7192 (6)&10.4483 (6)&0.0289s&PyTorch&\\
\hline\hline
input image & - &0.6750&0.6603&0.3103 &17.9391\\
\end{tabular}}
\caption{AIM 2020 Image Relighting Challenge Track 3 (Any-to-any relighting) results. The MPS, used to determine the final ranking, is computed following Eq.~\eqref{eq:MPS}. We also note that normalizing SSIM and (1-LPIPS) scores by the maximum in the track, for computing the MPS, does not affect the ranking.}
\label{table:T3}
\end{table}

\section{Track 1 methods}

\newcommand{\vidit}[1]{\includegraphics[width=0.31\linewidth]{#1}}
\begin{figure}
    \centering
    \begin{tabu}{ccc}
        \vidit{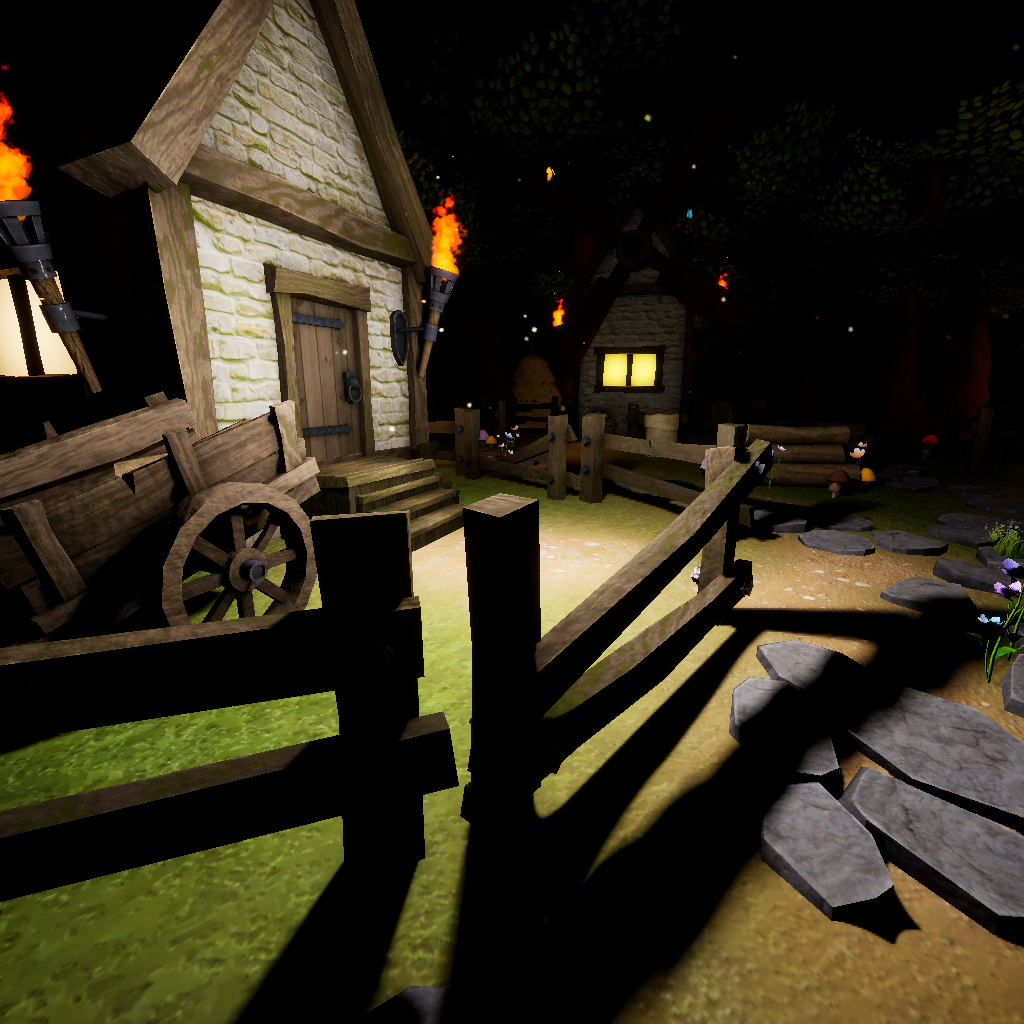} & 
        \vidit{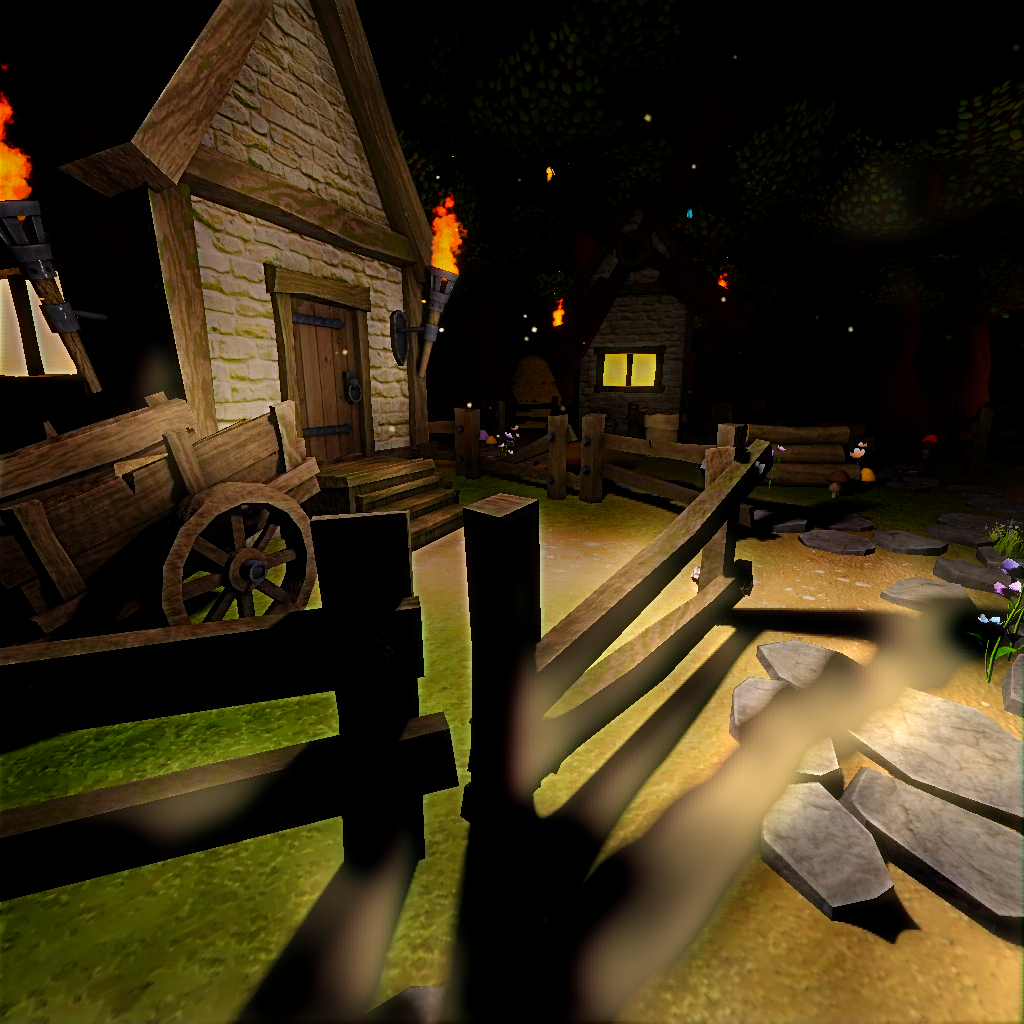} &
        \vidit{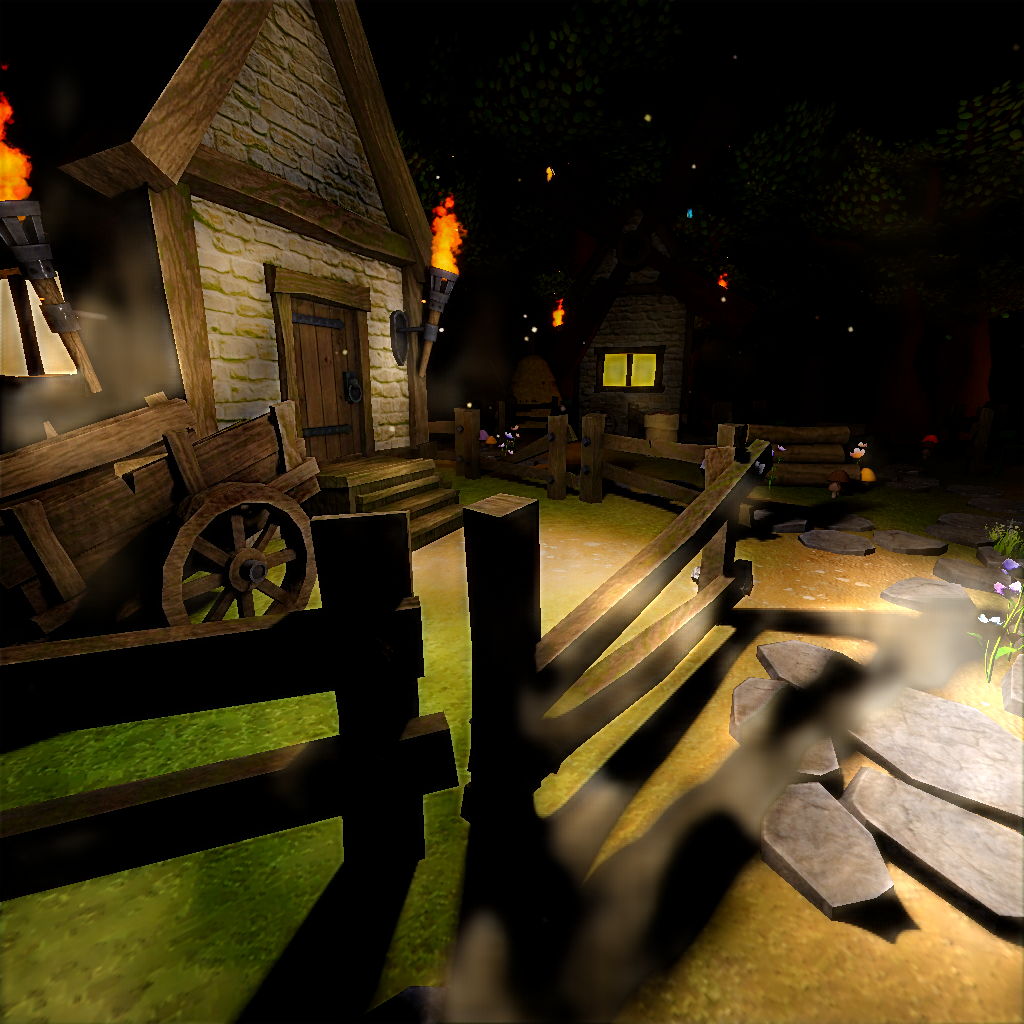} \\
        (a) Input & (b) CET\_SP (0.658)  & (c) CET\_CVLab (0.637)\\
        \vidit{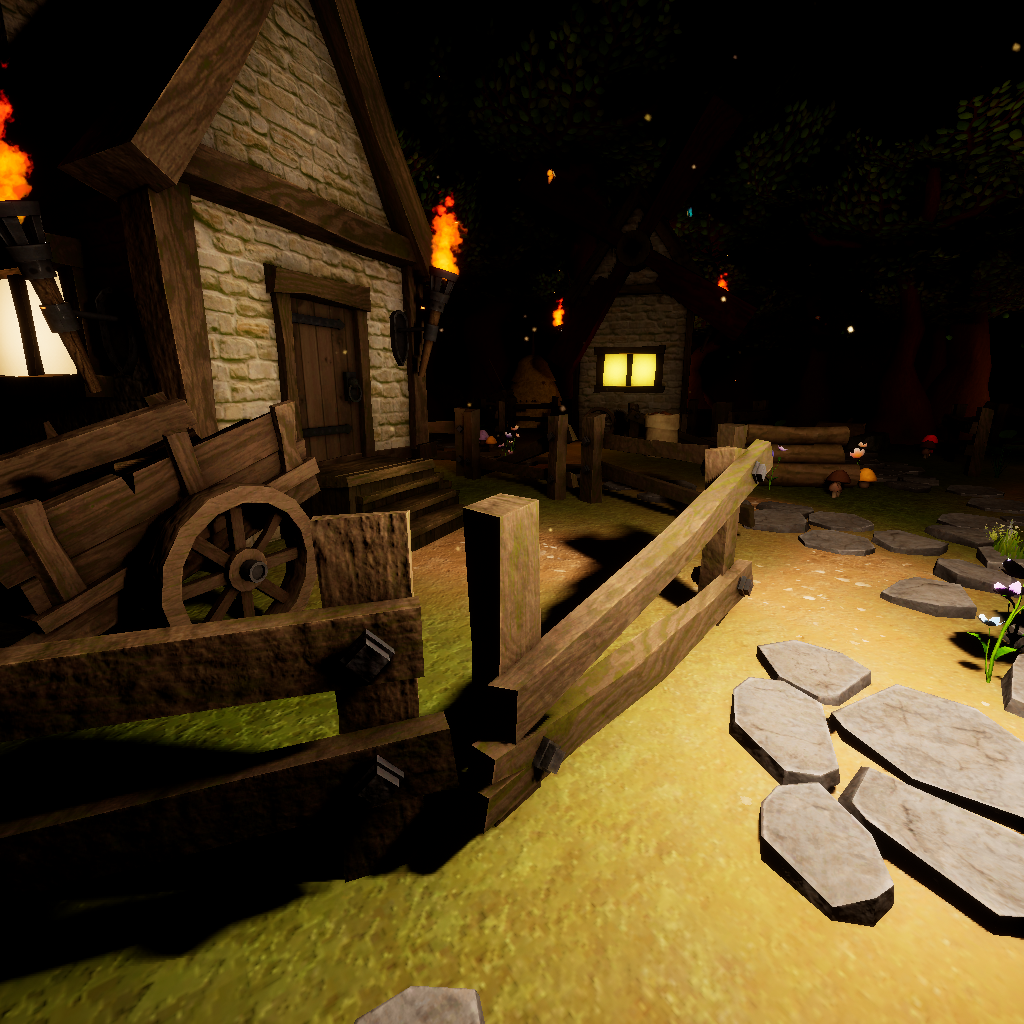}&
        \vidit{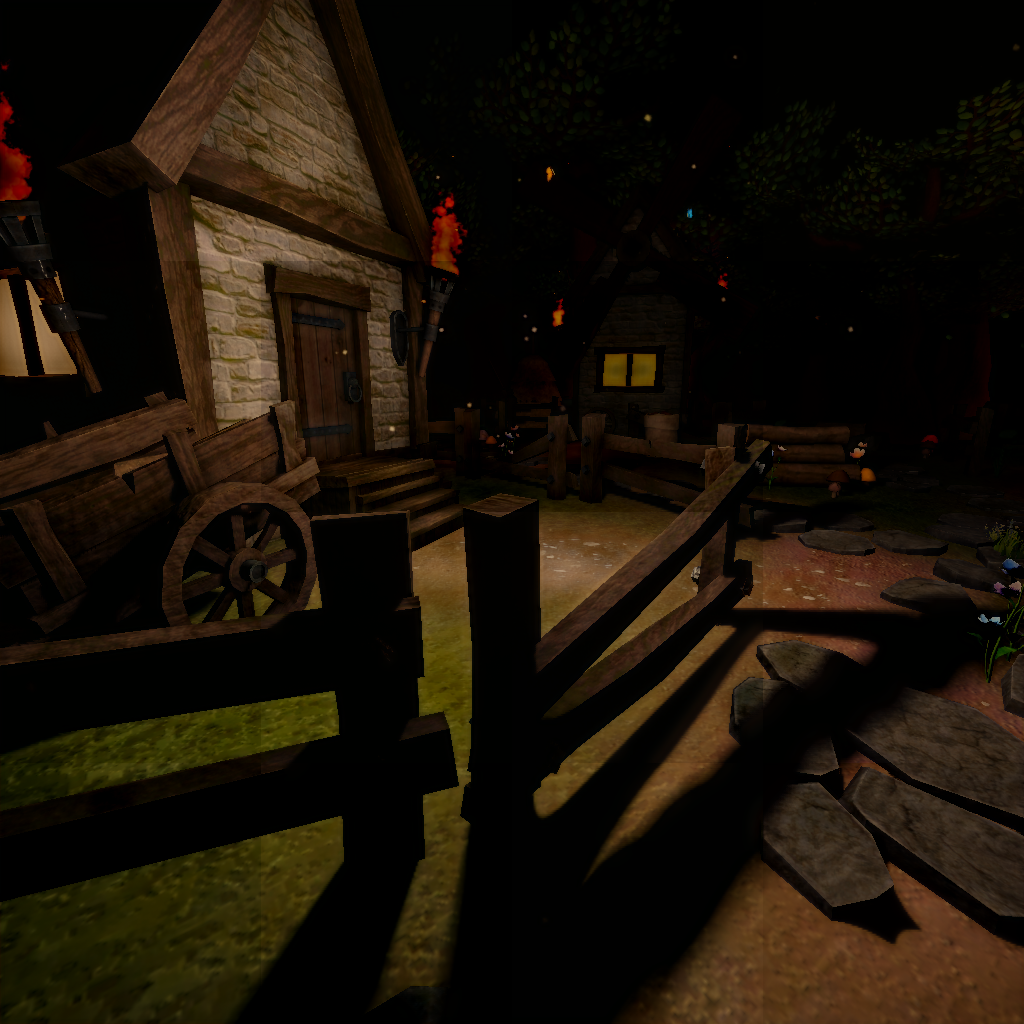}&
        \vidit{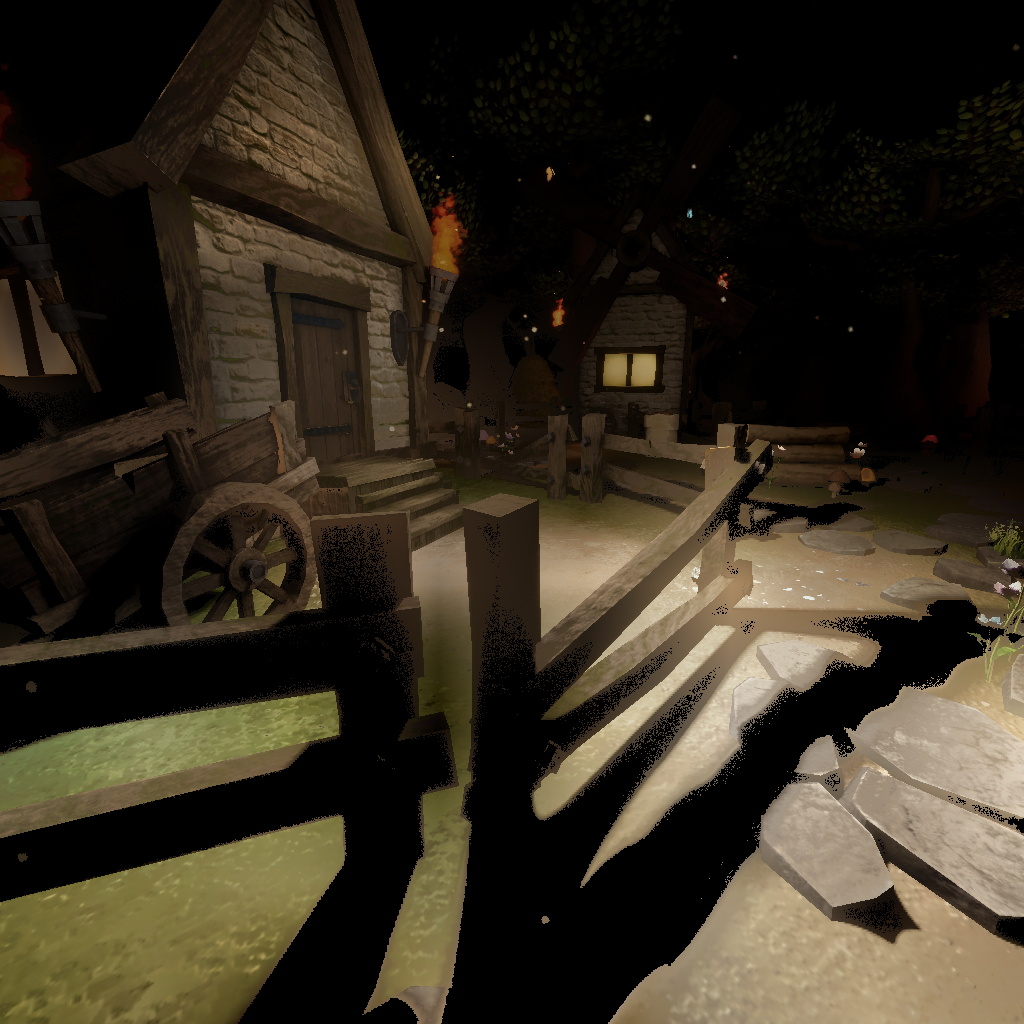}\\
        (d) Ground-truth & (e) lyl (0.610) & (f) YorkU (0.580)
    \end{tabu}
    \caption{Sample visual results from top submissions in track 1, with MPS scores. We observe that relighting previous shadows is the most difficult sub-task.}
    \label{fig:t1_visual}
\end{figure}
\begin{figure}
    \centering
    \begin{tabu}{ccc}
        \vidit{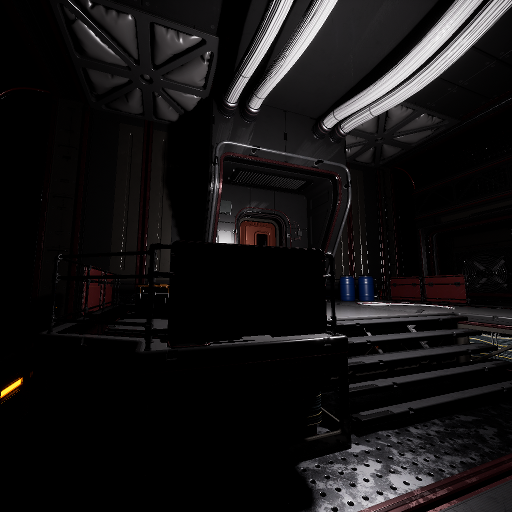} & 
        \vidit{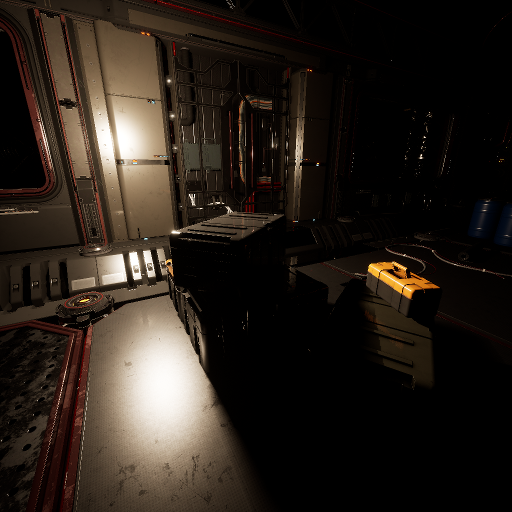} &
        \vidit{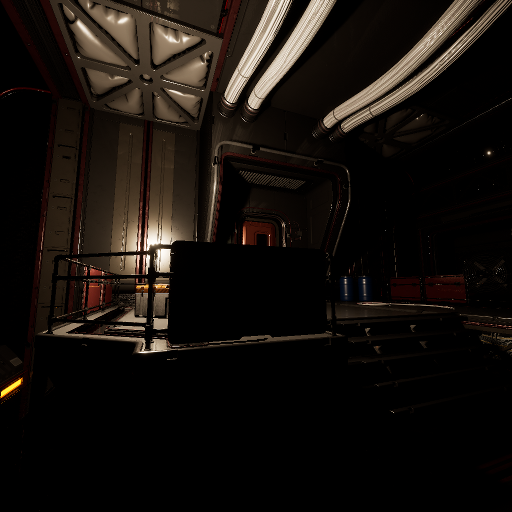} \\
        (a) Input & (b) Guide  & (c) Ground-truth\\
        \vidit{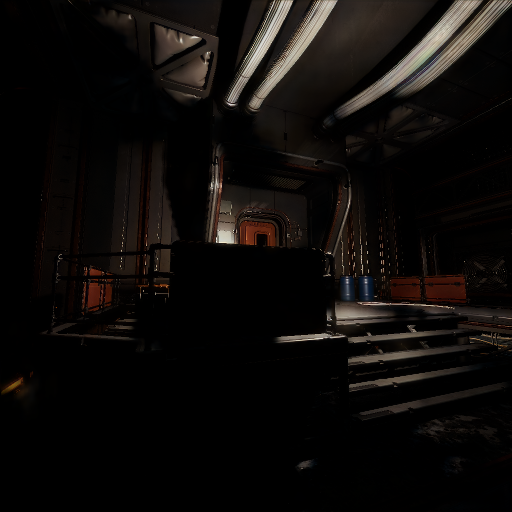}&
        \vidit{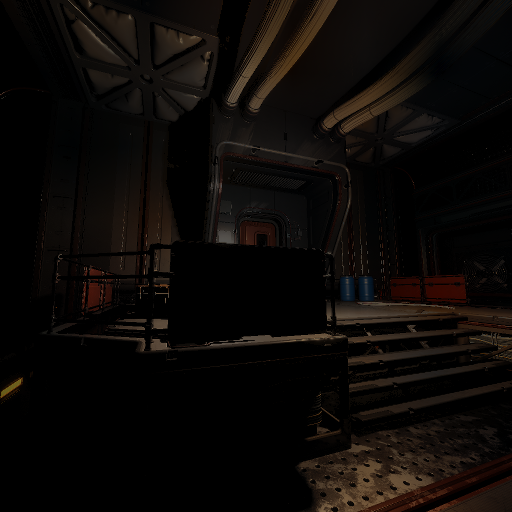}&
        \vidit{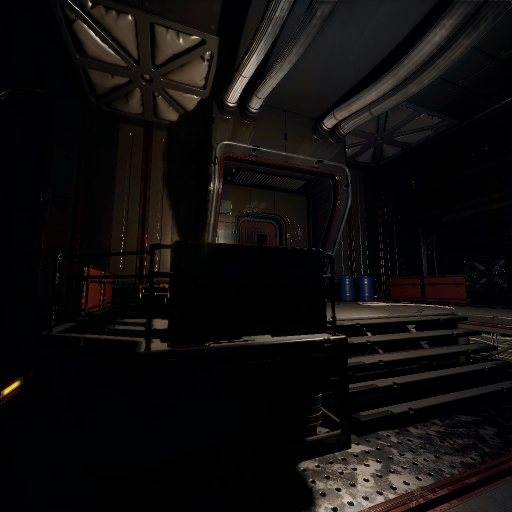}\\
        (d) NPU-CVPG (0.639) & (e) YorkU (0.628) & (f) IPCV\_IITM  (0.608)
    \end{tabu}
    \caption{Sample visual results from top submissions in track 3, with MPS scores.}
    \label{fig:t3_visual}
\end{figure}

\subsection{CET\_CVLab: Wavelet Decomposed RelightNet (WDRN)}
\begin{figure}[ht]
    \centering
    \includegraphics[scale=0.18]{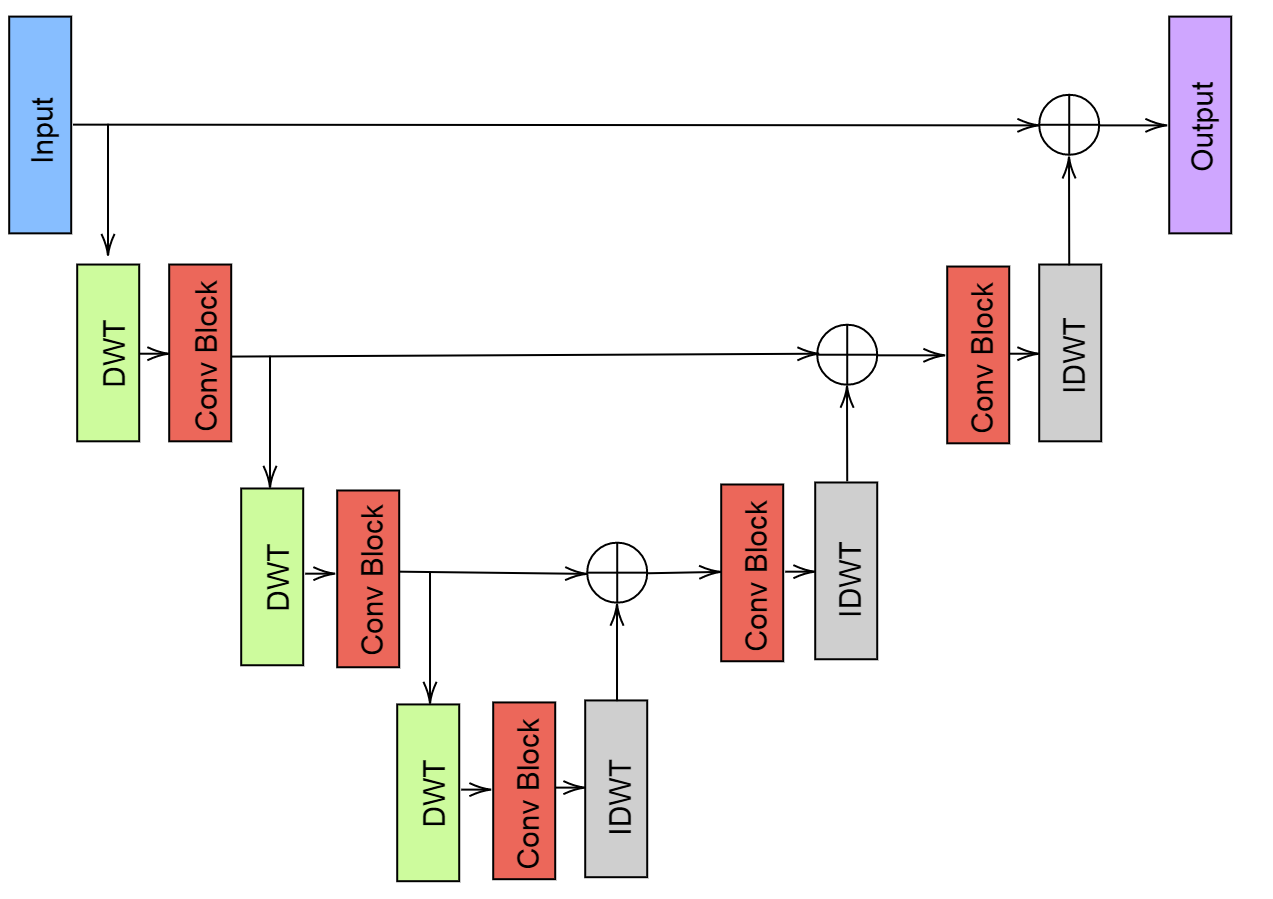}
    \caption{Architecture of the Wavelet Decomposed RelightNet (WDRN).}
    \label{fig:cvlab_wdrn}
\end{figure}

The architecture of the proposed Wavelet Decomposed RelightNet (WDRN)~\cite{puthussery2020wdrn} is shown in Fig.~\ref{fig:cvlab_wdrn}. The network structure used is similar to that of an encoder-decoder U-Net. The downsampling operation used in the contraction path is a discrete wavelet transform (DWT) based decomposition instead of a downsampling convolution or pooling. Similarly, in the expansion path, the inverse discrete wavelet transform (IDWT) is used instead of an upsampling convolution. In the wavelet based decomposition, the information from all channels is combined in the downsampling process such that there is minimal information loss when compared to that of a convolutional subsampling. For the given task, it can be deduced that the network must learn to re-calibrate the illumination gradient within the image. To this end, the network should be able to establish the relation between distant pixels. The proposed WDRN can achieve a high receptive field and hence establish this relation with the multi-scale wavelet decomposition. Also, this methodology is computationally efficient and is inspired by the multi-level wavelet-CNN (MWCNN) proposed by Liu \textit{et al.}~\cite{mwcnn}. The training loss used in this work is a weighted sum of the SSIM loss, MAE loss and a \textit{gray loss} (the gray loss term is used in the CET\_SP submission, and omitted in that of CET\_CVLab). Gray loss is the $\ell1$ distance between the grayscale version of the restored image and that of the ground-truth image.

\subsection{lyl: Coarse-to-Fine Relighting Net (CFRN)}\label{lyl}
\begin{figure}
    \centering
    \includegraphics[width=\textwidth]{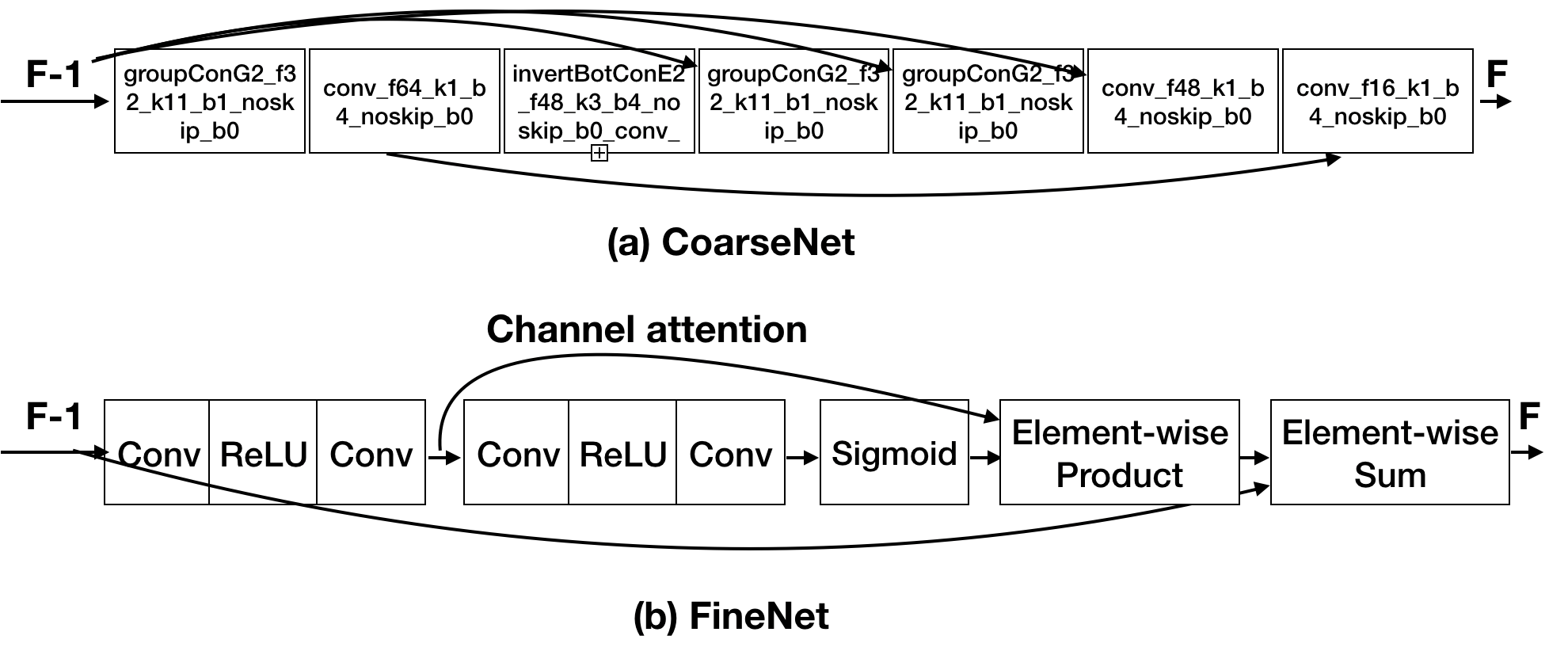}
    \caption{Architecture diagram of the Coarse-to-Fine Relighting Net (CFRN).}
    \label{fig:lyl_track1}
\end{figure}

The proposed Coarse-to-Fine Relighting Net (CFRN) is illustrated in Fig.~\ref{fig:lyl_track1}. The solution consists of two networks: (1) progressive coarse network and (2) a network merging the output of the coarse network, with channel attention, to correct the input in each level. Such a progressive process helps to achieve the principle for image relighting: high-level information is a good guide to obtain a better relit image. In the proposed method, there are three indispensable parts; (1) tying the loss at each level (2) using the FineNet structure and (3) providing a lower-level extracted feature input to ensure the availability of low-level information. To make full use of the training data, the team augments data in three ways; (1) scaling: randomly downscaling between [0.5,1.0], (2) rotation: randomly rotating the image by 90, 180, and 270 degrees, and (3) flipping: randomly flipping images horizontally or vertically with equal probability.

\subsection{YorkU: Norm-Relighting-U-Net (NRUNet)}\label{YorkU-one-to-one}
\begin{figure}
\centering
\includegraphics[width=\linewidth]{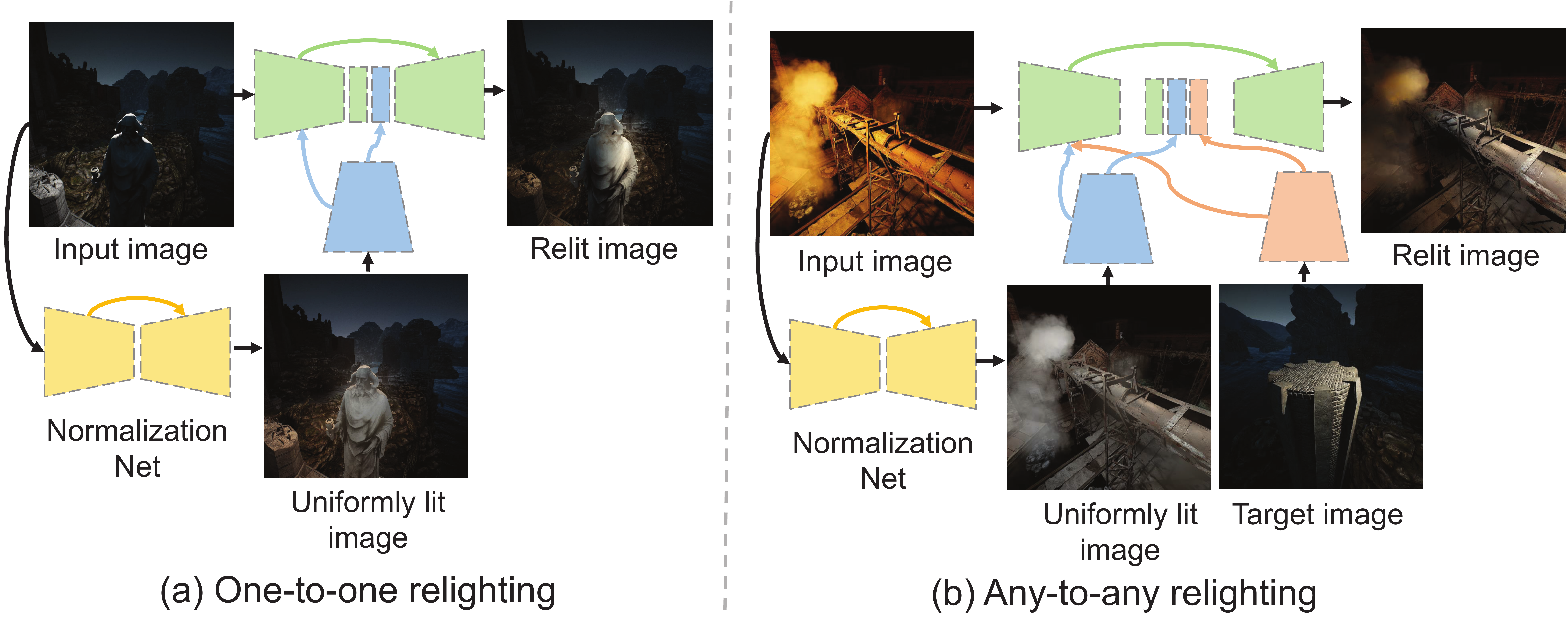}
\caption{Overview of YorkU team's NRUNet framework.}
\label{fig:YorkU_1}
\end{figure}

The method adopts a U-Net architecture \cite{ronneberger2015u} as the main backbone of the proposed framework. The solution consists of two networks: (1) the normalization network, which is responsible for producing uniformly-lit white-balanced images, and (2) the relighting network, which performs the one-to-one image relighting. An instance normalization~\cite{ulyanov2016instance} is applied after each stage in the encoder of the normalization network, while batch normalization is used for the encoder of the relighting network. The relighting network is fed the input image and the latent representations of the uniformly-lit image produced by the normalization network. The team uses the white-balance augmenter in~\cite{afifi2019else} to augment the training data. To produce the ground-truth of the normalization network, the team uses the training data provided for tracks 2 and 3, which include a set of images taken from each scene under different lighting directions. The team exploits their solution for the illumination settings estimation task (see Sec.~\ref{YorkU-ill}) to predict the target scene settings for the one-to-one mapping. Hence, the team increases the number of training images by including the training images provided for tracks 2 and 3. The team pre-trains the normalization network then fixes its weights and the entire framework is jointly trained. The training uses the Adam optimizer~\cite{kingma2014adam} with $\ell1$ loss. At inference, the team processes a resized version of the input image, then a guided up-sampling~\cite{chen2016bilateral} is applied to obtain the full-resolution image. The team ensembles the final results by utilizing their one-to-any framework (more details on the one-to-any framework in Sec.~\ref{YorkU-any-to-any}). To relight the image using the one-to-any framework, the team randomly selects six images with the predicted illumination settings of the current track to use them as targets. This procedure generates six relit images that are used along with the result image produced by the one-to-one framework to generate the final result. Fig.~\ref{fig:YorkU_1}-(a) shows an overview of the proposed one-to-one mapping framework. The source code for the three tracks is available at \url{https://github.com/mahmoudnafifi/image_relighting}.

\subsection{IPCV\_IITM: Deep Residual Network for Image Relighting (DRNIR)}
\begin{figure}
    \centering

    \includegraphics[width = 0.85\textwidth]{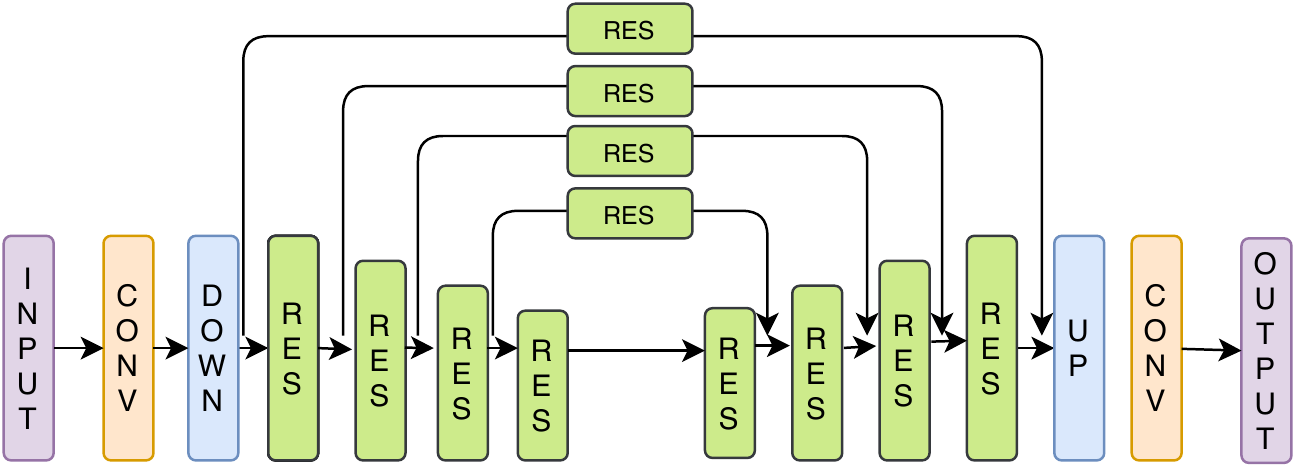}
    \caption{Diagram illustration of the DRNIR network architecture.}
    \label{fig:arch_ipcv_t1}
\end{figure}

Fig.~\ref{fig:arch_ipcv_t1} shows the structure of the proposed residual network with skip connections, based on the hourglass network~\cite{zhou2019deep}. The network has an encoder-decoder structure with skip connections~\cite{huang2017densely}. Residual blocks are used in the skip connections, and Batch-Norm and ReLU non-linearity in each of the blocks. The encoder features are concatenated with the decoder features of same level. The network takes the input image and directly produces the target image. The team converts the input RGB images to LAB for better processing. To reduce the memory consumption without harming the performance, the team uses a pixel-shuffle block \cite{shi2016real} to downsample the image. The network is first trained using the $\ell1$ loss, then fine-tuned with the MSE
loss. Note that experiments with adversarial loss did not lead to stable training. The learning rate of the Adam optimizer is 0.0001 with a decay cycle of 200 epochs, and a $512\times 512$ patch size for training. Data augmentation is used to make the network more robust.

\subsection{Other submitted solutions}
The DeepRelight team addresses the one-to-one relighting task by recovering the structure information of the scene, target illumination information, and renders the output with a GAN strategy~\cite{DRN2020}. Another solution makes use of two pairs of encoder-decoder networks, such that the encoding and decoding are illumination specific, and the learning is also supervised with discriminators. Transforming an image becomes equivalent to encoding it with the first encoder and decoding it with the second. Hertz tackle the problem using a multi-scale hierarchical network, the image is encoded at multiple resolutions and feature information is transferred from lower to higher levels to obtain the final transformation. Lastly, Image Lab~\cite{nathan2020lightnet} build on the multilevel hyper vision net~\cite{HyperVisionNet}, adding convolution block attention~\cite{WooCBAM} in their skip connections. Further details of each of these submitted solutions can be found in the supplementary material.

% \subsection{DeepRelight: Deep Relighting Network (DRN) for Image Light Source Manipulation}
% \begin{figure}
% \centering
% \includegraphics[width=.8\linewidth]{IMAGES/DRN.png}
% \caption{The architecture of Deep Relighting Network (DRN).}
% \label{fig:DRN}
% \end{figure}

% The proposed Deep Relighting Network (DRN) shown in Fig.~\ref{fig:DRN}) tackles the single image relighting task with three parts. First, it recovers the structure information of the scene. Second, it estimates the lighting effects (especially the shadows) for the target light source. Finally, a renderer combines the results and gives the final estimation. To improve the representation power of the auto-encoders, the down- and up-sampling processes are designed based on the back-projection theory~\cite{haris2018dbpn,DLN2020}. The manipulation of lighting effects needs to inpaint the shadows of the input and recast shadows following the target lighting. DRN uses the idea of generative adversarial learning~\cite{pix2pix,pix2pixHD} that measures the light effects though a shadow-region discriminator. %Both global and local information is essential for light source manipulation; global information for shadow region estimation and local information for detail reconstruction. 
% The renderer works through a multi-scale perception, which aggregates the global and local information for high-quality estimations.

\section{Track 2 methods}
\subsection{AiRiA\_CG: Dual Path Ensemble Network (DPENet)}
\begin{figure}
\centering
\includegraphics[width=.9\linewidth]{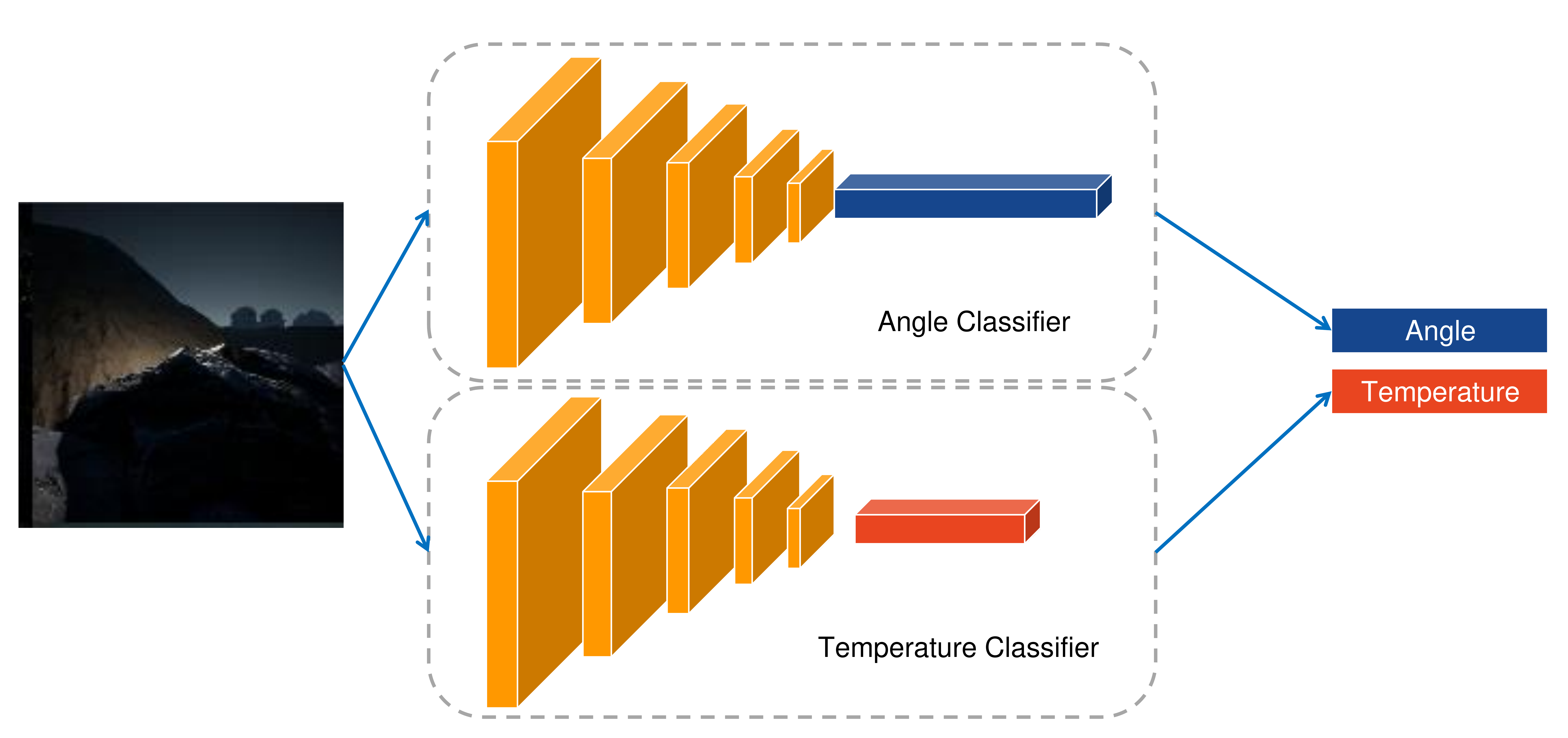}
\caption{The structure of Dual Path Ensemble Network (DPENet).}
\label{DPENet}
\end{figure}

The proposed DPENet has two sub-networks, one for angle prediction and one for temperature classification~\cite{liping2020aim}. The full DPENet is shown in Fig.~\ref{DPENet}. ResNeXt-101\_32$\times4$d~\cite{xie2017aggregated} is adopted for the angle prediction sub-network. The temperature classification sub-network is based on ResNet-50~\cite{he2016deep}. The two sub-networks are pre-trained on ImageNet~\cite{deng2009imagenet}. The solution adopts random flipping and random rotation for data augmentation. 
 
\subsection{YorkU: Illuminant-ResNet (I-ResNet)}\label{YorkU-ill}
\begin{figure}
\centering
\includegraphics[width=\linewidth]{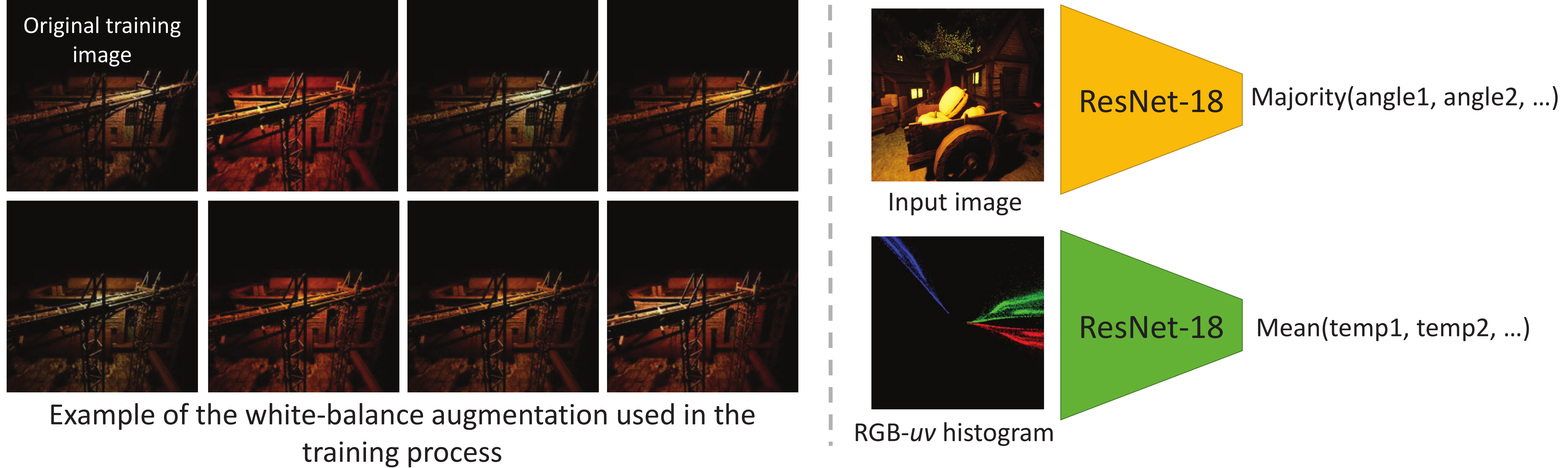}
\caption{Overview of the YorkU solution, with the white-balance augmentation~\cite{afifi2019else}.}
\label{fig:YorkU_2}
\end{figure}

The team treats the task as two independent classification tasks; (1) illuminant temperature classification and (2) illuminant angle classification. The team adopts the ResNet-18 model~\cite{he2016deep} trained on ImageNet~\cite{deng2009imagenet}. The last fully-connected layer is replaced with a new layer with $n$ neurons, where $n$ is the number of output classes for each task. The Adam optimizer~\cite{kingma2014adam} is used with cross entropy loss. For angle classification, the team applies the white-balance augmenter proposed in~\cite{afifi2019else} to augment the training data. For temperature classification, the team follows previous work~\cite{barron2015convolutional, afifi2019color, afifi2019sensor} that uses image histogram features instead of the 2D input image. Specifically, the team feeds the network with 2D RGB-$uv$ projected histogram features~\cite{afifi2019color, afifi2019sensor}, instead of the original training images. This histogram-based training, rather than image-based, improves the model's generalization. Fig.~\ref{fig:YorkU_2} shows an overview of the team's solution, including the white-balance augmentation process.

\subsection{Image Lab: Virtual Image Illumination Estimation (LightNet)}
\begin{figure}
\centering
\includegraphics[width=\linewidth]{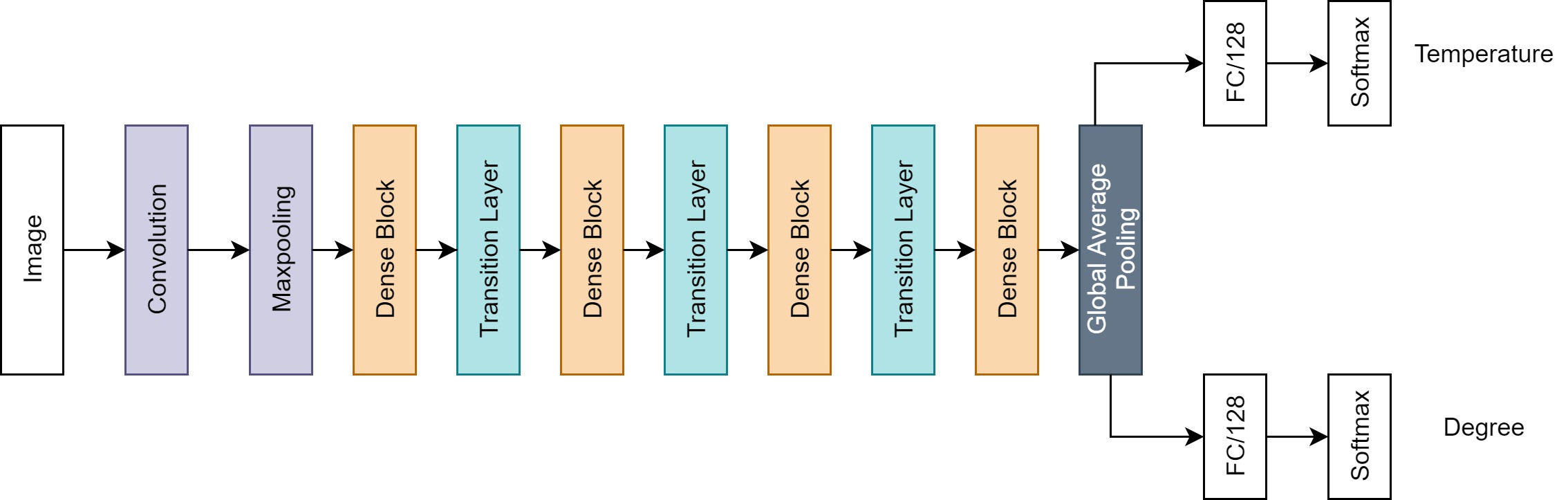}
\caption{Overview of the LightNet model's architecture.}
\label{fig:ImageLab_2}
\end{figure}

As shown in Fig.~\ref{fig:ImageLab_2}, the team adopts a Densenet~\cite{Densnet} architecture for the task. The team trains ten different pre-trained networks and also creates a custom network with selective blocks~\cite{Selective}. From these networks, the Densnet121 network achieves the best performance. %DenseNet has a property to avoid the vanishing gradient and reuse the feature properly.
DenseNet121 consists of fifty-eight dense blocks, followed by three transition blocks and three fully-connected layers. The global average pooling and fully connected layers are removed from the pre-trained network, and replaced with a new global average pooling and fully connected layers with a degree and temperature output layer. From the training dataset, the team creates a random splitting, with 67 percent of samples taken for training and the rest for validation. The training images are normalized to [0,1]. The Adam optimizer with a learning rate decaying from 0.001 to 0.00001 over 500 epochs is used for training the model with the categorical loss. Attention layers~\cite{WooCBAM} were tested in the development phase but did not yield any improvement.

\subsection{Other submitted solution}
The debut\_kele team proposes to use a single EfficientNet~\cite{tan2019efficientnet} backbone, pre-trained on ImageNet.
Further details of this submitted solution can be found in the supplementary material.

\section{Track 3 methods}
\subsection{NPU-CVPG: Self-Attention AutoEncoder (SA-AE)}
\begin{figure}
\includegraphics[width=\linewidth]{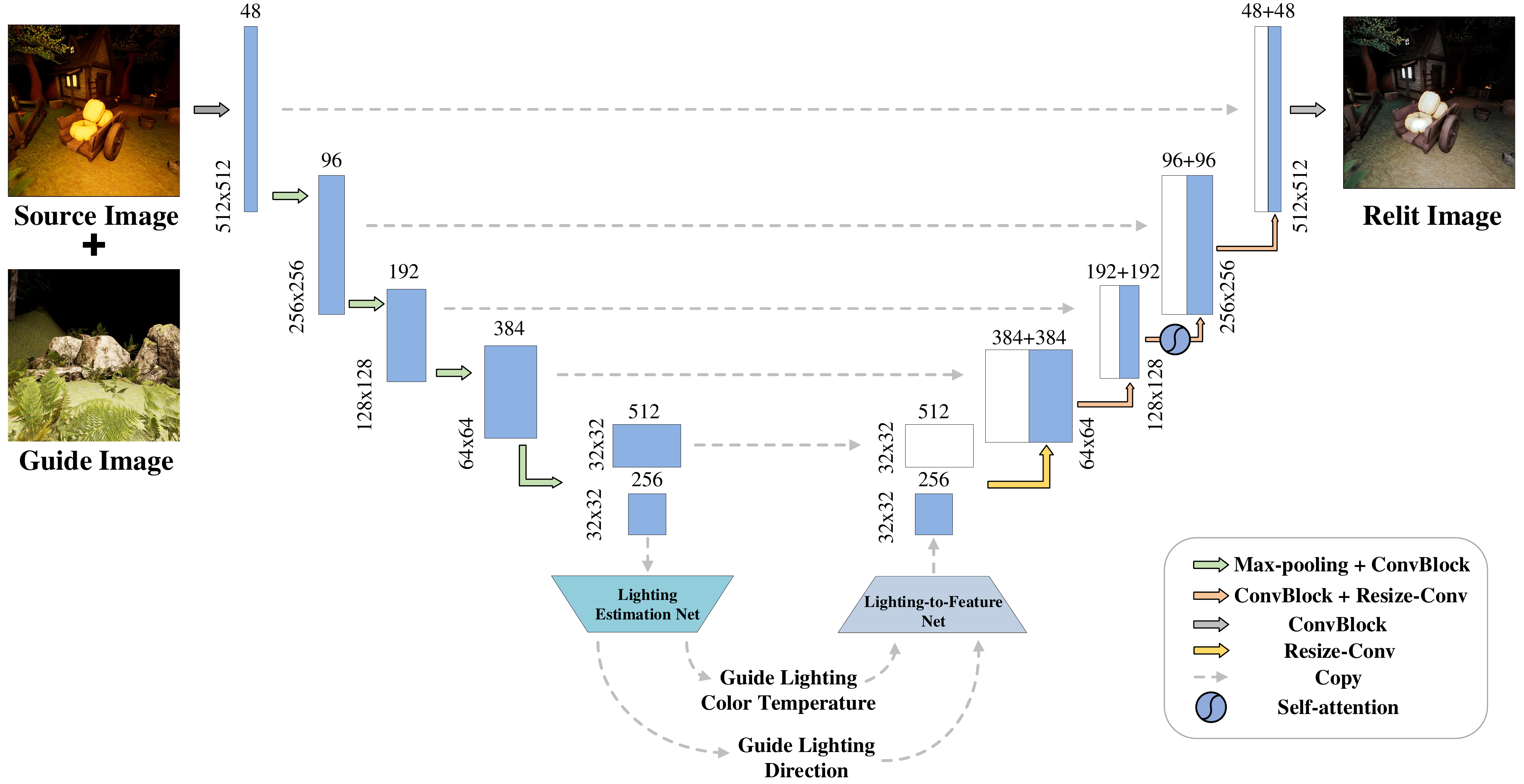}
\caption{Overview of the proposed SA-AE network.}
\label{fig:CVPG_Net}
\end{figure}

As shown in Fig.~\ref{fig:CVPG_Net}, the team presents the novel Self-Attention AutoEncoder (SA-AE)~\cite{hu2020SA-AE} model for generating a relit image from a source image to match the illumination settings of a guide image. In order to reduce the learning difficulty, the team adopts an implicit scene representation~\cite{zhou2019deep} learned by the encoder to render the relit images using the decoder. Based on the learned scene representation, an illumination estimation network is designed as a classier to predict the illumination settings of the guide image. A lighting-to-feature network is also designed to recover the corresponding implicit scene representation from the illumination settings, similar to the inverse of the illumination estimation process. In addition, a self-attention~\cite{zhang2019self} mechanism is introduced in the decoder to focus on the rendering of the regions requiring relighting in the source images.

\subsection{YorkU: Norm-Relighting-U-Net (NRUNet)}\label{YorkU-any-to-any}
As for the one-to-one mapping proposed (Sec.~\ref{YorkU-one-to-one}), the U-Net architecture~\cite{ronneberger2015u} is used as the main backbone of the any-to-any relighting framework, and two networks are used for normalization and relighting, as shown in Fig.~\ref{fig:YorkU_1}-(b). The relighting network is fed the input image, the latent representation of the guide image and the uniformly lit image produced by the normalization network. The team uses the white-balance augmentation \cite{afifi2019else} on the training data for the normalization network. The team trains two frameworks; one framework on $256\!\times\!256$ random patches and one on $256\!\times\!256$ resized images. The final result is generated by taking the mean of the two relit images and applying a guided up-sampling~\cite{chen2016bilateral}.

\subsection{IPCV\_IITM: Deep Residual Network for Image Relighting (DRNIR)}
\begin{figure}
    \centering
    \includegraphics[width = 0.85\textwidth]{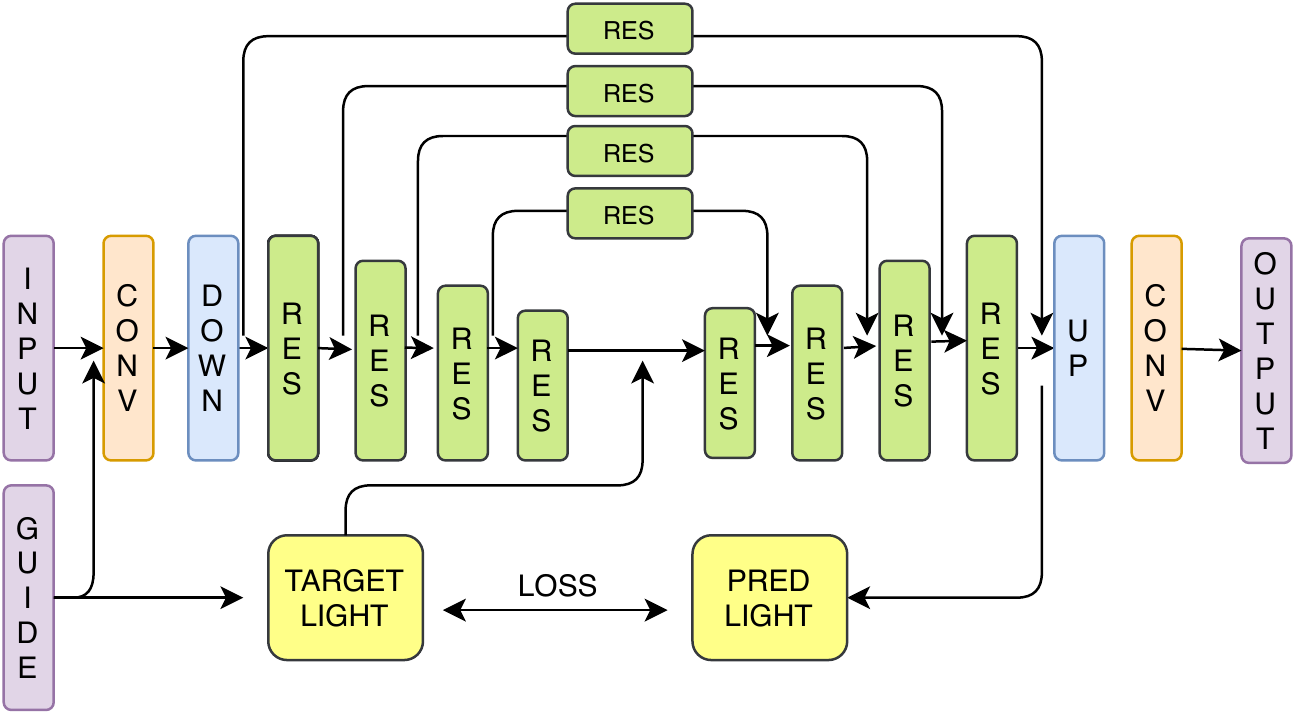}
    \caption{Network architecture of the DRNIR method.}
    \label{fig:arch_ipcv_t3}
\end{figure}

Fig.~\ref{fig:arch_ipcv_t3} shows the structure of the proposed residual network with skip connections, based on the hourglass network~\cite{zhou2019deep}. The network has an encoder-decoder structure similar to~\cite{huang2017densely}. The team also uses residual blocks in the skip connections. The encoder features are concatenated with the decoder features of the same level. Along with the input image, the network is given a guide image that is used in two places. First, both the input and the guide image are concatenated. Second, the team adds a separate loss to match the illumination properties between the guide image and the predicted image. A separate network predicts the illumination settings of an image, and is trained with the provided ground-truth labels. The team passes both the guide image and the predicted image through the network and minimizes the distance between intermediate feature representations. The feature representation of the guide image is further concatenated with the encoder output and fed to the decoder. The team converts the input RGB images to LAB for better processing. To reduce memory consumption, pixel-shuffle blocks \cite{shi2016real} are used as in track 1. %Our experiments with adversarial loss did not lead to stable training of our model.

\subsection{lyl: Coarse-to-Fine Relighting Net (CFRN)}
The proposed Coarse-to-Fine Relighting Net (CFRN) is shown in Fig.\ref{fig:lyl_track1}, as in track 1. Training is divided in two stages: incomplete training and full training.  %What's more, fine network was separate training with incomplete training. 
During an incomplete training, the fine network is trained with a batch size of 16 for 200 epochs. The Adam optimizer (${\beta_{1}=0.9}$,${\beta_{2}=0.999}$) is used to minimize the $\ell1$ loss between the generated relit images and the ground-truth. The learning rate is initialized to ${10^{4}}$ and kept unchanged. After the incomplete training with the fine network, the whole CFRN is fully trained. In each full training batch, the team randomly samples 64 patches for 20k epochs. 

\subsection{Other submitted solution}
The AiRiA\_CG team proposes a creative solution consisting of a dual encoder and single decoder~\cite{liping2020aim}. The input image is encoded, and so is the target image. However, the encoder of the target image is mirrored to match the decoder of the input image latent representation, and the feature layers of the former are thus transferred, layer by layer, to the decoder of the latter. This allows the illumination information to be transferred from the guide image to the input image during the decoding process. 
Further details of this submitted solution can be found in the supplementary material.

\appendix

\section*{Acknowledgements}
\addtocounter{footnote}{-2}
We thank all AIM 2020 sponsors: Huawei, MediaTek, NVIDIA, Qualcomm, Google and CVL, ETH Zurich (\url{https://data.vision.ee.ethz.ch/cvl/aim20/}). We also note that all tracks were supported by the CodaLab infrastructure (\url{https://competitions.codalab.org}).

%\newpage

\section{Teams and affiliations} \label{sec:teams}
% The teams who competed in the final test phase and have confirmed submissions are listed below in alphabetical order.\\

\noindent \textbf{AIM challenge organizers}\\
{\textit{Members}}: Majed  El  Helou,  Ruofan  Zhou,  Sabine  S\"usstrunk (\textit{\{majed.elhelou,  ruofan.zhou,sabine.susstrunk\}@epfl.ch}, EPFL, Switzerland), and Radu Timofte (\textit{radu.timofte@vision.ee.ethz.ch}, ETH Z\"urich, Switzerland).\\

\noindent \textbf{{-- AiRiA\_CG --}}\\
{\textit{Members}}: Yu Zhu(\textit{zhuyu.cv@gmail.com}), Liping Dong, Zhuolong Jiang, Chenghua Li, Cong Leng, Jian Cheng\\
{\textit{Affiliation}}: Nanjing Artificial Intelligence Chip Research, Institute of Automation, Chinese Academy of Sciences (AiRiA); MAICRO.

\noindent \textbf{{-- CET\_CVLab --}}\\
{\textit{Members}}: Densen Puthussery (\textit{puthusserydensen@gmail.com}), Hrishikesh P S, Melvin Kuriakose, Jiji C V\\
{\textit{Affiliation}}: College of Engineering, Trivandrum, India.

\noindent \textbf{{-- debut\_kele --}}\\
{\textit{Members}}: Kele Xu (\textit{kelele.xu@gmail.com}), Hengxing Cai, Yuzhong Liu \\
{\textit{Affiliation}}: National University of Defense Technology, China.

\noindent \textbf{{-- DeepRelight --}}\\
{\textit{Members}}: Li-Wen Wang$^1$ (\textit{liwen.wang@connect.polyu.hk}), Zhi-Song Liu$^{1,2}$, Chu-Tak Li$^1$, Wan-Chi Siu$^1$, Daniel P. K. Lun$^1$\\
{\textit{Affiliation}}: $^1$Department of Electronic and Information Engineering, The Hong Kong Polytechnic University, $^2$CS laboratory at the Ecole Polytechnique (Palaiseau).

\noindent \textbf{{-- Hertz --}}\\
{\textit{Members}}: Sourya Dipta Das$^1$ (\textit{dipta.juetce@gmail.com}), Nisarg A. Shah$^2$, \\Akashdeep Jassal$^3$\\
{\textit{Affiliation}}: $^1$Jadavpur University, Kolkata, India, $^2$Indian Institute of Technology Jodhpur, India, $^3$Punjab Engineering College (PEC), Chandigarh, India.

\noindent \textbf{{-- Image Lab --}}\\
{\textit{Members}}: Sabari Nathan$^1$ (\textit{sabarinathantce@gmail.com}),
M.Parisa Beham$^2$, R.Suganya$^3$\\
{\textit{Affiliation}}: $^1$Couger Inc, Tokyo, Japan, $^2$Sethu Institute of Technology, India, $^3$Thiagarajar College of Engineering, India.

\noindent \textbf{{-- IPCV\_IITM --}}\\
{\textit{Members}}: Maitreya Suin (\textit{maitreyasuin21@gmail.com}), Kuldeep Purohit, A. N. Rajagopalan\\
{\textit{Affiliation}}: Indian Institute of Technology Madras, India.

\noindent \textbf{{-- lyl --}}\\
{\textit{Members}}: Tongtong Zhao$^1$ (\textit{daitoutiere@gmail.com}), Shanshan Zhao$^2$\\
{\textit{Affiliation}}: $^1$Dalian Maritime University,$^2$ China Everbright Bank.

\noindent \textbf{{-- NPU-CVPG --}}\\
{\textit{Members}}: Zhongyun Hu (\textit{zy\_h@mail.nwpu.edu.cn}), Xin Huang, Yaning Li, Qing Wang\\
{\textit{Affiliation}}: Computer Vision and Computational Photography Group, School of Computer Science, Northwestern Polytechnical University.

\noindent \textbf{{-- RGETH --}}\\
{\textit{Members}}: George Chogovadze (\textit{chogeorg@student.ethz.ch}), R\'{e}mi Pautrat\\
{\textit{Affiliation}}: ETH Zurich, Switzerland.

\noindent \textbf{{-- YorkU --}}\\
{\textit{Members}}: Mahmoud Afifi (\textit{mafifi@eecs.yorku.ca}), Michael S. Brown \\
{\textit{Affiliation}}: EECS, York University, Toronto, ON, Canada.

% \clearpage
% ---- Bibliography ----
%
% BibTeX users should specify bibliography style 'splncs04'.
% References will then be sorted and formatted in the correct style.
%
\bibliographystyle{splncs04}
\bibliography{egbib}

\begin{thebibliography}{10}
\providecommand{\url}[1]{\texttt{#1}}
\providecommand{\urlprefix}{URL }
\providecommand{\doi}[1]{https://doi.org/#1}

\bibitem{afifi2019sensor}
Afifi, M., Brown, M.S.: Sensor-independent illumination estimation for
  {D}{N}{N} models. In: British Machine Vision Conference (BMVC) (2019)

\bibitem{afifi2019else}
Afifi, M., Brown, M.S.: What else can fool deep learning? addressing color
  constancy errors on deep neural network performance. In: IEEE International
  Conference on Computer Vision (ICCV). pp. 243--252 (2019)

\bibitem{afifi2019color}
Afifi, M., Price, B., Cohen, S., Brown, M.S.: When color constancy goes wrong:
  Correcting improperly white-balanced images. In: IEEE Conference on Computer
  Vision and Pattern Recognition (CVPR). pp. 1535--1544 (2019)

\bibitem{barron2015convolutional}
Barron, J.T.: Convolutional color constancy. In: IEEE International Conference
  on Computer Vision (ICCV). pp. 379--387 (2015)

\bibitem{barron2012color}
Barron, J.T., Malik, J.: Color constancy, intrinsic images, and shape
  estimation. In: European Conference on Computer Vision (ECCV). pp. 57--70
  (2012)

\bibitem{bell2014intrinsic}
Bell, S., Bala, K., Snavely, N.: Intrinsic images in the wild. ACM Transactions
  on Graphics (TOG)  \textbf{33}(4), ~159 (2014)

\bibitem{bousseau2009user}
Bousseau, A., Paris, S., Durand, F.: User-assisted intrinsic images. In: ACM
  SIGGRAPH Asia, pp. 1--10 (2009)

\bibitem{burton1987color}
Burton, G.J., Moorhead, I.R.: Color and spatial structure in natural scenes.
  Applied Optics  \textbf{26}(1),  157--170 (1987)

\bibitem{cabon2020virtual}
Cabon, Y., Murray, N., Humenberger, M.: Virtual kitti 2. arXiv preprint
  arXiv:2001.10773  (2020)

\bibitem{chen2016bilateral}
Chen, J., Adams, A., Wadhwa, N., Hasinoff, S.W.: Bilateral guided upsampling.
  ACM Transactions on Graphics (TOG)  \textbf{35}(6), ~1--8 (2016)

\bibitem{deng2009imagenet}
Deng, J., Dong, W., Socher, R., Li, L.J., Li, K., Fei-Fei, L.: Image{N}et: A
  large-scale hierarchical image database. In: IEEE Conference on Computer
  Vision and Pattern Recognition (CVPR). pp. 248--255 (2009)

\bibitem{dherse2020scene}
Dherse, A.P., Everaert, M.N., Gwizda{\l}a, J.J.: Scene relighting with
  illumination estimation in the latent space on an encoder-decoder scheme.
  arXiv preprint arXiv:2006.02333  (2020)

\bibitem{liping2020aim}
Dong, L., Jiang, Z., Li, C.: An ensemble neural network for scene relighting
  with light classification. In: Proceedings of the European Conference on
  Computer Vision Workshops (ECCVW) (2020)

\bibitem{HyperVisionNet}
D.Sabarinathan, Beham, M., Roomi, S.: Moire image restoration using multi level
  hyper vision net. Image and Video Processing  (2020), arXiv:2004.08541

\bibitem{elhelou2018aam}
El~Helou, M., D{\"u}mbgen, F., S{\"u}sstrunk, S.: {AAM}: An assessment metric
  of axial chromatic aberration. In: IEEE International Conference on Image
  Processing (ICIP). pp. 2486--2490 (2018)

\bibitem{elhelou2020vidit}
El~Helou, M., Zhou, R., Barthas, J., S{\"u}sstrunk, S.: {VIDIT}: Virtual image
  dataset for illumination transfer. arXiv preprint arXiv:2005.05460  (2020)

\bibitem{elhelou2020aim_relighting}
El~Helou, M., Zhou, R., Süsstrunk, S., Timofte, R., et~al.: {AIM 2020}: Scene
  relighting and illumination estimation challenge. In: European Conference on
  Computer Vision Workshops (2020)

\bibitem{finlayson2004intrinsic}
Finlayson, G.D., Drew, M.S., Lu, C.: Intrinsic images by entropy minimization.
  In: European Conference on Computer Vision (ECCV). pp. 582--595 (2004)

\bibitem{fuoli2020aim_VXSR}
Fuoli, D., Huang, Z., Gu, S., Timofte, R., et~al.: {AIM 2020} challenge on
  video extreme super-resolution: Methods and results. In: European Conference
  on Computer Vision Workshops (2020)

\bibitem{he2016deep}
He, K., Zhang, X., Ren, S., Sun, J.: Deep residual learning for image
  recognition. In: IEEE Conference on Computer Vision and Pattern Recognition
  (CVPR). pp. 770--778 (2016)

\bibitem{hu2020SA-AE}
Hu, Z., Huang, X., Li, Y., Wang, Q.: {SA-AE} for any-to-any relighting. In:
  Proceedings of the European Conference on Computer Vision Workshops (ECCVW)
  (2020)

\bibitem{Densnet}
{Huang}, G., {Liu}, Z., {Van Der Maaten}, L., {Weinberger}, K.Q.: Densely
  connected convolutional networks. In: IEEE Conference on Computer Vision and
  Pattern Recognition (CVPR). pp. 2261--2269 (2017)

\bibitem{huang2017densely}
Huang, G., Liu, Z., Van Der~Maaten, L., Weinberger, K.Q.: Densely connected
  convolutional networks. In: IEEE conference on computer vision and pattern
  recognition (CVPR). pp. 4700--4708 (2017)

\bibitem{ignatov2020aim_ISP}
Ignatov, A., Timofte, R., et~al.: {AIM 2020} challenge on learned image signal
  processing pipeline. In: European Conference on Computer Vision Workshops
  (2020)

\bibitem{ignatov2020aim_bokeh}
Ignatov, A., Timofte, R., et~al.: {AIM 2020} challenge on rendering realistic
  bokeh. In: European Conference on Computer Vision Workshops (2020)

\bibitem{kingma2014adam}
Kingma, D.P., Ba, J.: Adam: A method for stochastic optimization. arXiv
  preprint arXiv:1412.6980  (2014)

\bibitem{kovacs2017shading}
Kovacs, B., Bell, S., Snavely, N., Bala, K.: Shading annotations in the wild.
  In: IEEE Conference on Computer Vision and Pattern Recognition (CVPR). pp.
  6998--7007 (2017)

\bibitem{Selective}
{Li}, X., {Wang}, W., {Hu}, X., {Yang}, J.: Selective kernel networks. In: IEEE
  Conference on Computer Vision and Pattern Recognition (CVPR). pp. 510--519
  (2019)

\bibitem{li2018learning}
Li, Z., Snavely, N.: Learning intrinsic image decomposition from watching the
  world. In: IEEE Conference on Computer Vision and Pattern Recognition (CVPR).
  pp. 9039--9048 (2018)

\bibitem{mwcnn}
Liu, P., Zhang, H., Zhang, K., Lin, L., Zuo, W.: Multi-level wavelet-{CNN} for
  image restoration. In: IEEE Conference on Computer Vision and Pattern
  Recognition (CVPR) Workshops. pp. 773--782 (2018)

\bibitem{llanos2020simultaneous}
Llanos, B., Yang, Y.H.: Simultaneous demosaicing and chromatic aberration
  correction through spectral reconstruction. In: IEEE Conference on Computer
  and Robot Vision (CRV). pp. 17--24 (2020)

\bibitem{matsushita2004illumination}
Matsushita, Y., Nishino, K., Ikeuchi, K., Sakauchi, M.: Illumination
  normalization with time-dependent intrinsic images for video surveillance.
  Transactions on Pattern Analysis and Machine Intelligence  \textbf{26}(10),
  1336--1347 (2004)

\bibitem{murmann2019dataset}
Murmann, L., Gharbi, M., Aittala, M., Durand, F.: A dataset of
  multi-illumination images in the wild. In: IEEE International Conference on
  Computer Vision (ICCV). pp. 4080--4089 (2019)

\bibitem{nagano2019deep}
Nagano, K., Luo, H., Wang, Z., Seo, J., Xing, J., Hu, L., Wei, L., Li, H.: Deep
  face normalization. ACM Transactions on Graphics (TOG)  \textbf{38}(6), ~183
  (2019)

\bibitem{nathan2020lightnet}
Nathan, D.S., Beham, M.P.: {LightNet:} deep learning based illumination
  estimation from virtual images. In: European Conference on Computer Vision
  Workshops (2020)

\bibitem{ntavelis2020aim_inpainting}
Ntavelis, E., Romero, A., Bigdeli, S.A., Timofte, R., et~al.: {AIM 2020}
  challenge on image extreme inpainting. In: European Conference on Computer
  Vision Workshops (2020)

\bibitem{puthussery2020wdrn}
Puthussery, D., P~S, H., Kuriakose, M., C.~V., J.: {WDRN}: A wavelet decomposed
  relightnet for image relighting. In: European Conference on Computer Vision
  Workshops (2020)

\bibitem{ronneberger2015u}
Ronneberger, O., Fischer, P., Brox, T.: U-{N}et: Convolutional networks for
  biomedical image segmentation. In: MICCAI. pp. 234--241 (2015)

\bibitem{shen2011intrinsic}
Shen, J., Yang, X., Jia, Y., Li, X.: Intrinsic images using optimization. In:
  IEEE Conference on Computer Vision and Pattern Recognition (CVPR). pp.
  3481--3487 (2011)

\bibitem{shi2016real}
Shi, W., Caballero, J., Husz{\'a}r, F., Totz, J., Aitken, A.P., Bishop, R.,
  Rueckert, D., Wang, Z.: Real-time single image and video super-resolution
  using an efficient sub-pixel convolutional neural network. In: IEEE
  conference on computer vision and pattern recognition (CVPR). pp. 1874--1883
  (2016)

\bibitem{son2020aim_VTSR}
Son, S., Lee, J., Nah, S., Timofte, R., Lee, K.M., et~al.: {AIM 2020} challenge
  on video temporal super-resolution. In: European Conference on Computer
  Vision Workshops (2020)

\bibitem{sun2019single}
Sun, T., Barron, J.T., Tsai, Y.T., Xu, Z., Yu, X., Fyffe, G., Rhemann, C.,
  Busch, J., Debevec, P., Ramamoorthi, R.: Single image portrait relighting.
  ACM Transactions on Graphics (TOG)  \textbf{38}(4), ~79 (2019)

\bibitem{tan2019efficientnet}
Tan, M., Le, Q.V.: Efficientnet: Rethinking model scaling for convolutional
  neural networks. arXiv preprint arXiv:1905.11946  (2019)

\bibitem{tappen2003recovering}
Tappen, M.F., Freeman, W.T., Adelson, E.H.: Recovering intrinsic images from a
  single image. In: Advances in neural information processing systems. pp.
  1367--1374 (2003)

\bibitem{torralba2003statistics}
Torralba, A., Oliva, A.: Statistics of natural image categories. Network:
  Computation in Neural Systems  \textbf{14}(3),  391--412 (2003)

\bibitem{ulyanov2016instance}
Ulyanov, D., Vedaldi, A., Lempitsky, V.: Instance normalization: The missing
  ingredient for fast stylization. arXiv preprint arXiv:1607.08022  (2016)

\bibitem{DRN2020}
Wang, L.W., Siu, W.C., Liu, Z.S., Li, C.T., Lun, D.P.: Deep relighting networks
  for image light source manipulation. In: Proceedings of the European
  Conference on Computer Vision Workshops (ECCVW) (2020)

\bibitem{wang2019underexposed}
Wang, R., Zhang, Q., Fu, C.W., Shen, X., Zheng, W.S., Jia, J.: Underexposed
  photo enhancement using deep illumination estimation. In: IEEE Conference on
  Computer Vision and Pattern Recognition (CVPR). pp. 6849--6857 (2019)

\bibitem{wang2004image}
Wang, Z., Bovik, A.C., Sheikh, H.R., Simoncelli, E.P.: Image quality
  assessment: from error visibility to structural similarity. IEEE Transactions
  on Image Processing  \textbf{13}(4),  600--612 (2004)

\bibitem{wei2020aim_realSR}
Wei, P., Lu, H., Timofte, R., Lin, L., Zuo, W., et~al.: {AIM 2020} challenge on
  real image super-resolution. In: European Conference on Computer Vision
  Workshops (2020)

\bibitem{weiss2001deriving}
Weiss, Y.: Deriving intrinsic images from image sequences. In: IEEE
  International Conference on Computer Vision (ICCV). vol.~2, pp. 68--75 (2001)

\bibitem{WooCBAM}
Woo, S., Park, J., Lee, J.Y., Kweon, I.S.: {CBAM} convolutional block attention
  module. Proceedings of the European Conference on Computer Vision (ECCV) pp.
  1--17 (2018)

\bibitem{xie2017aggregated}
Xie, S., Girshick, R., Doll{\'a}r, P., Tu, Z., He, K.: Aggregated residual
  transformations for deep neural networks. In: IEEE conference on computer
  vision and pattern recognition (CVPR). pp. 1492--1500 (2017)

\bibitem{xu2018deep}
Xu, Z., Sunkavalli, K., Hadap, S., Ramamoorthi, R.: Deep image-based relighting
  from optimal sparse samples. ACM Transactions on Graphics (TOG)
  \textbf{37}(4), ~126 (2018)

\bibitem{zhang2019self}
Zhang, H., Goodfellow, I., Metaxas, D., Odena, A.: Self-attention generative
  adversarial networks. In: International Conference on Machine Learning
  (ICML). pp. 7354--7363 (2019)

\bibitem{zhang2020aim_efficientSR}
Zhang, K., Danelljan, M., Li, Y., Timofte, R., et~al.: {AIM 2020} challenge on
  efficient super-resolution: Methods and results. In: European Conference on
  Computer Vision Workshops (2020)

\bibitem{zhang2018unreasonable}
Zhang, R., Isola, P., Efros, A.A., Shechtman, E., Wang, O.: The unreasonable
  effectiveness of deep features as a perceptual metric. In: IEEE Conference on
  Computer Vision and Pattern Recognition (CVPR). pp. 586--595 (2018)

\bibitem{zhao2020modified}
Zhao, J., Hou, Y., Liu, Z., Xie, H., Liu, S.: Modified color {CCD} moir{\'e}
  method and its application in optical distortion correction. Precision
  Engineering  (2020)

\bibitem{zhou2019deep}
Zhou, H., Hadap, S., Sunkavalli, K., Jacobs, D.W.: Deep single-image portrait
  relighting. In: IEEE International Conference on Computer Vision (ICCV). pp.
  7194--7202 (2019)

\end{thebibliography}
\end{document}